\documentclass[10pt,twocolumn,letterpaper]{article}

\usepackage{cvpr}              % To produce the CAMERA-READY version
\usepackage[inkscapelatex=false]{svg}
\usepackage{bm} 
\usepackage{multirow}
\usepackage{algorithm}  
\usepackage{algpseudocode}  
\usepackage{amsmath}  
\usepackage{tabularx}
\usepackage{array} % 提供 \fontsize 和 \selectfont 命令
\usepackage{booktabs} % 提供更好的表格排版样式
\usepackage{pifont}
\usepackage{wasysym}
\usepackage{graphicx}
\usepackage{subcaption}
\usepackage[accsupp]{axessibility}  % Improves PDF readability for those with disabilities.
\definecolor{cvprblue}{rgb}{0.21,0.49,0.74}
\usepackage[pagebackref,breaklinks,colorlinks,allcolors=cvprblue]{hyperref}

\def\paperID{13612} % *** Enter the Paper ID here
\def\confName{CVPR}
\def\confYear{2025}

\title{Generalized Gaussian Entropy Model for Point Cloud Attribute Compression with Dynamic Likelihood Intervals}

\author{
Changhao Peng, Yuqi Ye, Wei Gao*\\
School of Electronic and Computer Engineering, Shenzhen Graduate Scool, Peking University\\
{\tt\small pch@stu.pku.edu.cn, yeyuqi0303@stu.pku.edu.cn, gaowei262@pku.edu.cn}
}

\begin{document}
\maketitle
\begin{abstract}
Gaussian and Laplacian entropy models are proved effective in learned point cloud attribute compression, as they assist in arithmetic coding of latents. However, we demonstrate through experiments that there is still unutilized information in entropy parameters estimated by neural networks in current methods, which can be used for more accurate probability estimation. Thus we introduce generalized Gaussian entropy model, which controls the tail shape through shape parameter to more accurately estimate the probability of latents. Meanwhile, to the best of our knowledge, existing methods use fixed likelihood intervals for each integer during arithmetic coding, which limits model performance. We propose Mean Error Discriminator (MED) to determine whether the entropy parameter estimation is accurate and then dynamically adjust likelihood intervals. Experiments show that our method significantly improves rate-distortion (RD) performance on three VAE-based models for point cloud attribute compression, and our method can be applied to other compression tasks, such as image and video compression.

% The ABSTRACT is to be in fully justified italicized text, at the top of the left-hand column, below the author and affiliation information.
% Use the word ``Abstract'' as the title, in 12-point Times, boldface type, centered relative to the column, initially capitalized.
% The abstract is to be in 10-point, single-spaced type.
% Leave two blank lines after the Abstract, then begin the main text.
% Look at previous \confName abstracts to get a feel for style and length.
\end{abstract}    
\section{Introduction}

In recent years, point clouds have become an important type of 3D visual data, with widespread applications in virtual reality \cite{wirth2019pointatme}, autonomous driving \cite{yue2018lidar}, and various other fields. With the massive generation of point cloud, there is an urgent need for efficient compression methods. Point clouds consist of two main aspects: geometry information and attribute information. Geometry information refers to the 3D coordinates of the points, while attribute information includes various details such as color and reflectance. Current geometry compression methods, such as the traditional coding algorithms used in G-PCC \cite{li2024mpeggpcc} and V-PCC \cite{li2024mpegvpcc}, as well as deep learning-based methods like SparsePCGC \cite{wang2022sparse}, OctAttention \cite{fu2022octattention} and EHEM \cite{song2023efficient}, have achieved efficient 
\begin{figure}[ht]
\centering
\includegraphics[width=\columnwidth]{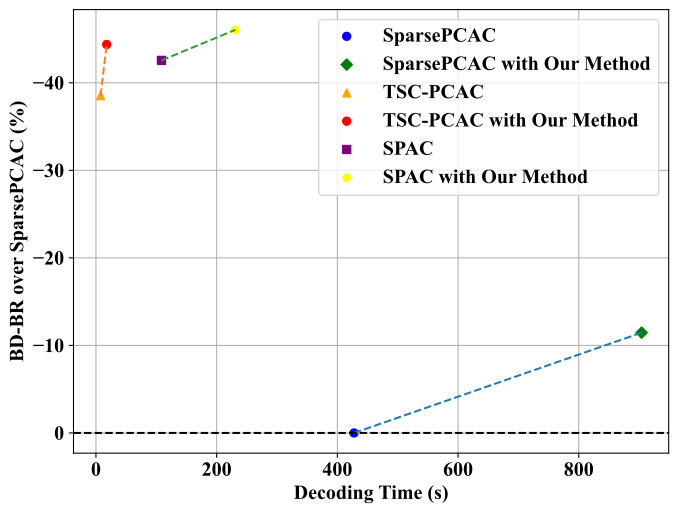}
% \captionsetup{font={small,bf,stretch=1.25}, justification=raggedright}
 \captionsetup{font={small}, singlelinecheck=off}
\caption{Rate-speed comparison on 8iVFBv2 \cite{8iVFBv2} and MVUB \cite{MVUB}. Left-top is better.}
\label{fig:head}
\end{figure}
\begin{figure}[ht]
\centering
\includegraphics[width=\columnwidth]{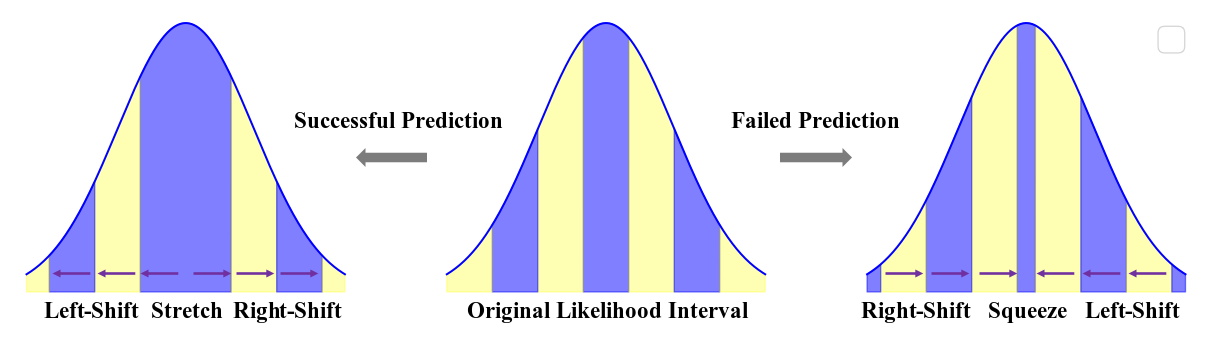}
% \captionsetup{font={small,bf,stretch=1.25}, justification=raggedright}
 \captionsetup{font={small}, singlelinecheck=off}
\caption{Dynamic likelihood intervals. When the prediction of mean is inaccurate, squeeze the likelihood interval of $\lfloor\mu\rceil$ so that the remaining integer likelihood intervals can shift towards the center of the entropy model, thereby increasing the likelihood. On the contrary, when the prediction is accurate, the likelihood interval of $\lfloor\mu\rceil$ is stretched.}
\label{fig:dli}
\end{figure}
compression of point cloud geometry. However, for attribute compression, the achievable compression ratios with current methods are still far below those for geometry compression. Therefore, our focus is on point cloud attribute compression based on lossless geometry.

Variational autoencoders (VAEs) \cite{kingma2013auto} have played an important role in deep learning-based compression frameworks. Based on VAE, Ballé et al. \cite{balle2018variational} propose using hyperpriors to generate multivariate Gaussian distributions for arithmetic coding of latents. Later, in point cloud attribute compression, Gaussian or Laplacian distributions determined by the mean and scale, have been used as entropy models for latents \cite{minnen2018joint,he2021checkerboard,minnen2020channel}. In fact, we demonstrate through experiments that these entropy model parameters can be utilized to further improve the performance of the model.

On the other hand, we need to transform continuous entropy models into discrete probability tables for arithmetic coding \cite{said1999introduction}, where each integer corresponds to a probability. Previous work \cite{balle2018variational} directly uses the half neighborhood of integers as likelihood intervals to calculate discrete probability tables. Our experiment shows that this fixed likelihood interval limits the performance of the model.

Based on the above discussion, we propose:
\begin{itemize}
\item[$\bullet$]Using generalized Gaussian distribution which is more expressive as the entropy model to provide more accurate probability estimation of the latents.
\item[$\bullet$]Determining the accuracy of the current entropy model based on mean, scale and decoded latents, then dynamically adjusting the likelihood intervals for more efficient arithmetic coding.
\item [$\bullet$]Developing a two-step training strategy, as our method does not make significant changes to the original structure of the model except adding a few modules, allowing it to be well applied to current VAE-based models.
\end{itemize}

\section{Related Work}
\subsection{Learned Image Compression}
Learned image compression is a highly developed field, and its methods are of significant reference value for the compression of point cloud attributes. Ballé et al. \cite{balle2018variational} are the first to propose an image compression method based on hyperpriors, which has become the foundational framework for subsequent image compression techniques. Building on the hyperprior model, Minnen et al. \cite{minnen2018joint} introduce a context module that allows for more accurate estimation of entropy parameters. However, the serial computation required by the context module substantially increases encoding and decoding times. He et al. \cite{he2021checkerboard} propose a checkerboard model that utilizes context from anchor regions to assist in the encoding of non-anchor regions, reducing the encoding and decoding time with almost no performance loss. Cheng et al. \cite{cheng2020learned} suggest replacing Gaussian distributions with mixture Gaussian distributions as the entropy model. Minnen et al. \cite{minnen2020channel} propose leveraging context between channels to assist in the encoding process. Recent work \cite{he2022elic,jiang2023mlic,jiang2023mlic++} has focused on maximizing the use of contextual information while maintaining the parallelism of the context modules.

\begin{figure}[ht]
  % 创建一个minipage，用于放置图片
  \begin{minipage}{0.5\columnwidth}
    \centering
    \includegraphics[width=\textwidth]{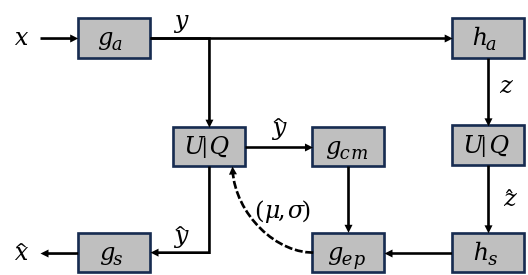}
    % \captionof{figure}{This is a figure caption}
  \end{minipage}%
  % 使用hfill来在图片和表格之间添加水平填充
  % \hfill
  % 创建另一个minipage，用于放置表格
  \begin{minipage}{0.5\columnwidth}
   \scriptsize
    \centering
    \begin{tabular}{cc}
      \textbf{Component} & \textbf{Symbol} \\
      \hline
      Input/Output Point Cloud& $(\bm{x},\bm{\hat{x}})$ \\
      Autoencoder& $(g_a,g_s)$ \\
      Hyper Autoencoder& $(h_a,h_s)$ \\
      (Quantized) Latents& $(\bm{y},\bm{\hat{y}})$ \\
      (Quantized) Hyper Latents& $(\bm{z},\bm{\hat{z}})$ \\
      Context Model&$g_{cm}$\\
      Entropy Parameters& $g_{ep}$\\
      Mean and Scale& $(\bm{\mu},\bm{\sigma})$

    \end{tabular}
  \end{minipage}
  \captionsetup{font={small}, singlelinecheck=off}
  \caption{Operational diagrams: Variational compression model with hyperprior and ontext.}
\label{fig:vae}
\end{figure}

\subsection{Point Cloud Attribute Compression}
In traditional point cloud attribute compression methods, the main research focus is on how to optimize prediction and transformation. Song et al. \cite{song2022fine} propose using different prediction methods for point clouds with different features, Shao et al. \cite{shao2017attribute} propose using graph transform instead of discrete cosine transform (DCT), and Peng et al. \cite{peng2024laplacian} propose learning an optimal Laplacian matrix through convex optimization to achieve better graph transform.

In recent years, the performance of deep learning-based methods has gradually surpassed traditional methods. Wang et al. \cite{wang2022sparse} propose the first learned point cloud attribute compression framework based on context and hyperpriors. Recent work \cite{wang2024versatile} aims at how to more efficiently utilize context. Guo et al. \cite{guo2024tsc} propose using context between channels to increase the parallelism of context modules and combining sparse convolution and transformer to better capture local and global information. Mao et al. \cite{mao2024spac} propose layering point clouds to more fully utilize context achieves SOTA RD performance. Zhang et al. \cite{zhang2023scalable,ding2022carnet} also propose using neural networks to enhance traditional methods, which compensates for the shortcomings of traditional methods' insufficient predictive ability.

\section{Preliminary}

\subsection{Variational Compression Model with Hyperprior and Context}

The framework of the variational compression model with hyperprior and context is shown in Figure \ref{fig:vae}. $\bm{y}$ is the latent representation after encoding, while $\bm{z}$ is the hyper latent for estimating the entropy model of $\bm{y}$. During training, quantization is replaced by adding uniform noise from $\mathcal{U}(-\frac{1}{2},\frac{1}{2})$ to $\bm{y}$ or $\bm{z}$. The entropy model for the decoding part of $\bm{\hat{y}}$ is jointly estimated by $\bm{\hat{z}}$ and the context obtained from the decoded part, while $\bm{\hat{z}}$ is compressed by a factorized entropy model. The model uses rate-distortion cost as the loss function for end-to-end training:
\begin{equation}
\begin{split}
R + \lambda \cdot D &= E_{\bm{x} \sim p_{\bm{x}}} \left[ -\log_{2} p_{\bm{\hat{y}}|\bm{\hat{z}}}(\bm{\hat{y}}|\bm{\hat{z}}) - \log_{2} p_{\bm{\hat{z}}}(\bm{\hat{z}}) \right] \\
&+ \lambda \cdot E_{\bm{x} \sim p_{\bm{x}}} [d(\bm{x}, \bm{\hat{x}})]
\end{split}
\end{equation}
where entropy is used to estimate the bitrate of $\bm{\hat{y}}$ and $\bm{\hat{z}}$, and $d(\bm{x}, \bm{\hat{x}})$ is the distortion of the reconstructed point cloud, usually represented by MSE or MS-SSIM \cite{wang2004image}.

In fact, $\bm{\hat{y}}$ occupies the vast majority of the total bitrate (usually more than 95\%), of which the probability is modeled by Gaussian likelihoods:
\begin{equation}
    p_{\bm{\hat{y}}|\bm{\hat{z}}}(\bm{\hat{y}}|\bm{\hat{z}})=\prod_i \left[ \mathcal{N}(\mu_i, \sigma_i^2) * \mathcal{U}\left(-\frac{1}{2}, \frac{1}{2}\right) \right] (\hat{y}_i).
    \label{gaussian likelihood}
\end{equation}
The Gaussian distribution in Equation \ref{gaussian likelihood} could be replaced with mixture Gaussian distribution or Laplacian distribution.

\subsection{Generalized Gaussian Distribution}
\begin{figure}[htbp]
\begin{minipage}{0.5\columnwidth}
\centering
        \includegraphics[width=\textwidth]{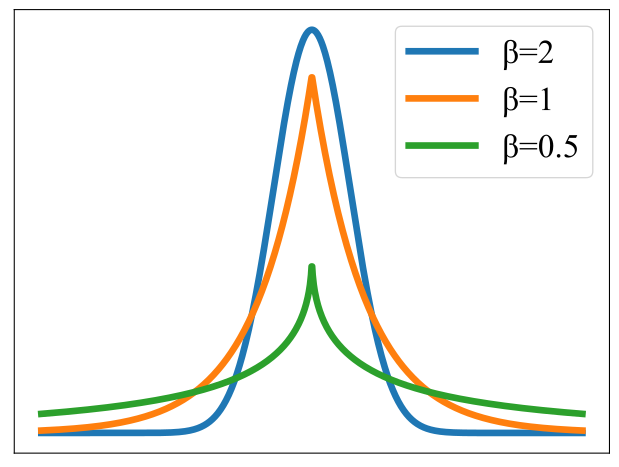}
        \subcaption{}
        \label{ggd}
    \end{minipage}
    \hfill
\begin{minipage}{0.473\columnwidth}
\centering
        \includegraphics[width=\textwidth]{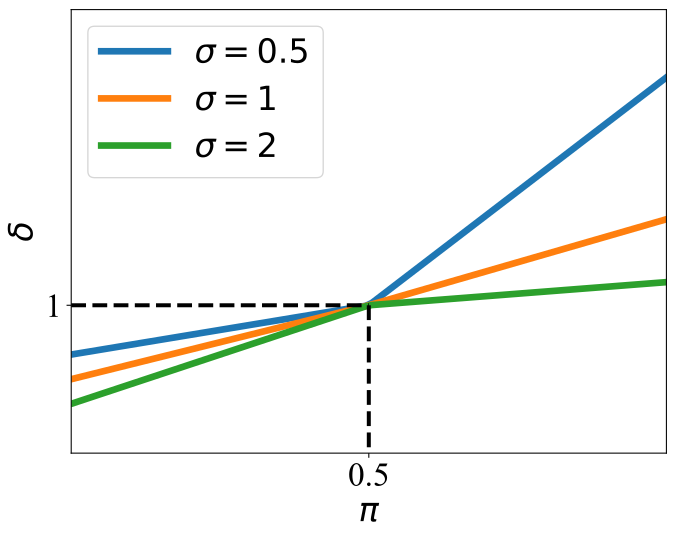}
        \subcaption{}
        \label{delta}
    \end{minipage}
    \hfill
\begin{minipage}{1\columnwidth}
\centering
        \includegraphics[width=\textwidth]{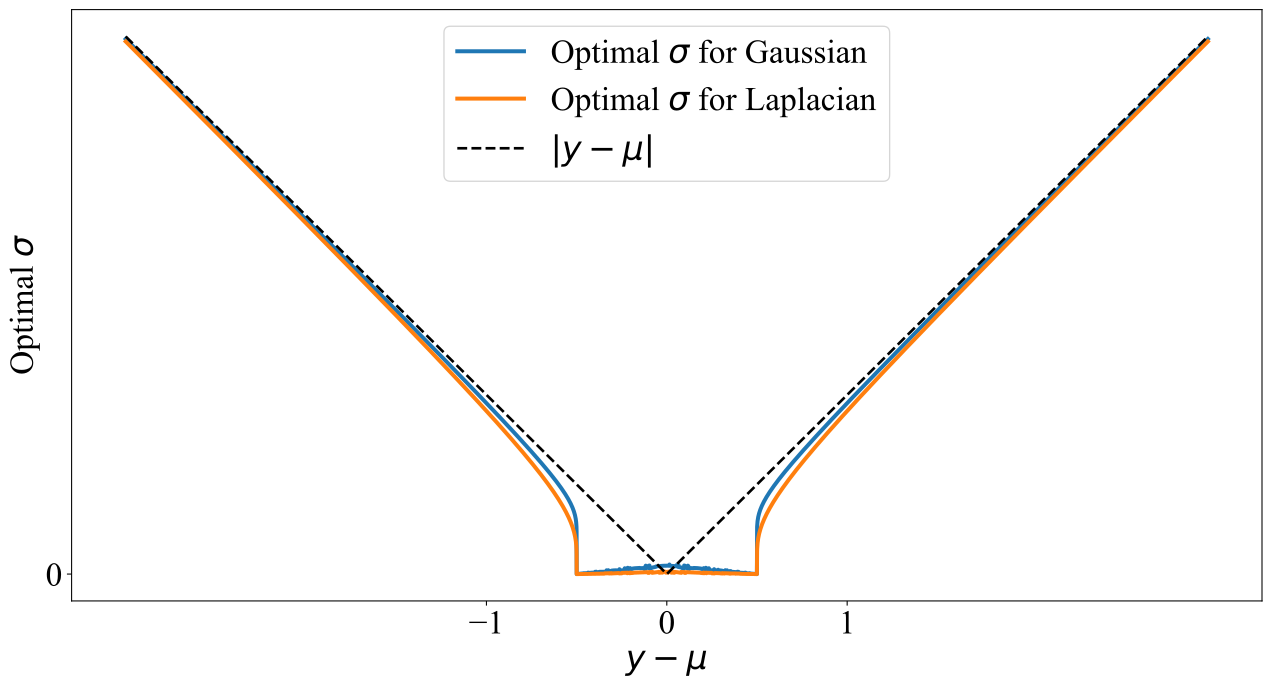}
        \subcaption{}
        \label{y-mu}
    \end{minipage}
\captionsetup{font={small}, singlelinecheck=off}
  \caption{Function diagram: (a) Generalized Gaussian distribution with different shape parameter $\beta$ while $\mu=0$. (b) The relationship between likelihood interval radius $\delta$ of $\lfloor\mu\rceil$ and $\pi=P(\hat{y}=\lfloor\mu\rceil)$ ,which is detailed discussed in Section \ref{Dynamic Likelihood Interval}. (c) Optimal scale $\sigma$ for $y-\mu$ using Gaussian distribution and Laplacian distribution as the entropy model respectively.}
  \label{fig:3}
\end{figure}

If the random variable $X\sim\mathcal{G}(\mu,\sigma,\beta)$,then
\begin{equation}
    P(X=x)=\frac{\beta}{2 \sigma \Gamma\left(\frac{1}{\beta}\right)} e^{-\frac{|x - \mu|^\beta}{\sigma^\beta}}
    \label{pdf}
\end{equation}
where
\begin{equation}
    \Gamma(z) = \int_0^\infty t^{z-1} e^{-t} \, dt.
\end{equation}
Compared to the Gaussian distribution, generalized Gaussian distribution \cite{bustin2018analytical} has an additional shape parameter $\beta$ to control whether the distribution has a heavy tail or a light tail, as shown in Figure \ref{ggd}.

It is worth noting that the generalized Gaussian distribution degenerates into a Laplacian distribution when $\beta=1$, or a Gaussian distribution when $\beta=2$.

\section{Explore a More Accurate Likelihood}

\subsection{The Relationship Between Latent and Entropy Parameters}
Obviously, the closer $\bm{\mu}$ is to $\bm{\hat{y}}$ in Equation \ref{gaussian likelihood}, the greater the likelihood and the lower the bitrate, which can be written as:
\begin{equation}
    \arg\max_{\bm{\mu}} p_{\bm{\hat{y}}|\bm{\hat{z}}}(\bm{\hat{y}}|\bm{\hat{z}})=\bm{\hat{y}},
\end{equation}
thus we can consider $\bm{\mu}$ as the predicted value of $\bm{\hat{y}}$. Next, we analyze the optimal value of the scale parameter $\bm{\sigma}$ by fixing $\bm{\mu}$ and $\bm{\hat{y}}$:
\begin{equation}
    \arg\max_{\bm{\sigma}} p_{\bm{\hat{y}}|\bm{\hat{z}}}(\bm{\hat{y}}|\bm{\hat{z}}).
\end{equation}
We can conclude that the $i$-th element of $\bm{\sigma}$ satisfies:
\begin{equation}
    \sigma_{i}=\arg\max_{\sigma} \left[ \mathcal{N}(\mu_i, \sigma^2) * \mathcal{U}\left(-\frac{1}{2}, \frac{1}{2}\right) \right] (\hat{y}_i).
\end{equation}
Since
\begin{equation}
    \mathcal{N}(\mu_i, \sigma^2)(\hat{y}_i)=\frac{1}{\sqrt{2\pi\sigma^2}} e^{-\frac{(\hat{y}_i-\mu_i)^2}{2\sigma^2}},
\end{equation}
for situations with accurate predictions ($\mu_i$ and $\hat{y}_i$ are close), the bitrate is relatively low. Therefore, we consider the case of inaccurate predictions where $\mu_i$ and $\hat{y}_i$ have a noticeable difference. To simplify the calculations in this situation, we make the following approximations:
\begin{equation}
    \begin{split}
         f(\hat{y}_i,\mu_i,\sigma):&=\frac{1}{\sqrt{2\pi}\sigma} e^{-\frac{(\hat{y}_i-\mu_i)^2}{2\sigma^2}} \\
         &\approx \frac{1}{\sqrt{2\pi}\sigma} \int_{\hat{y}_i-\mu_i-\frac{1}{2}}^{\hat{y}_i-\mu_i+\frac{1}{2}} e^{-\frac{t^2}{2\sigma^2}} \, dt\\
         &=\int_{\hat{y}_i-\frac{1}{2}}^{\hat{y}_i+\frac{1}{2}} \frac{1}{\sqrt{2\pi\sigma^2}} e^{-\frac{(t-\mu_i)^2}{2\sigma^2}} \, dt \\
         &=\left[ \mathcal{N}(\mu_i, \sigma^2) * \mathcal{U}\left(-\frac{1}{2}, \frac{1}{2}\right) \right] (\hat{y}_i).
    \end{split}
\end{equation}
Then find $\sigma^{opt}$ by the partial derivative:
\begin{equation}
\begin{split}
    \frac{\partial f}{\partial \sigma^{opt}}&=\frac{1}{\sqrt{2\pi}}e^{-\frac{(\hat{y}_i-\mu_i)^2}{2(\sigma^{opt})^2}} \left [ \frac{(\hat{y}_i-\mu_i)^2}{(\sigma^{opt})^4}-\frac{1}{(\sigma^{opt})^2}\right ] \\
    &=0,
\end{split}
\end{equation}
so the optimal $\sigma_i$ for $(\hat{y}_i,\mu_i)$ can be approximated by $\sigma^{opt}=|\hat{y}_i-\mu_i|$. Moreover, we can get the same result if the Gaussian distribution in Equation \ref{gaussian likelihood} is replaced with a mixture Gaussian distribution (fix the mixture weights $\bm{w}$) or Laplacian distribution. To visually show the relationship between $\sigma^{opt}(y,\mu)$ and $|y-\mu|$, we numerically compute $\sigma^{opt}(y,\mu)$ and present the results in Figure \ref{y-mu}. 

In general, the optimal $\bm{\sigma}$ can be approximated as:
\begin{equation}
\begin{split}
    \arg\max_{\bm{\sigma}} p_{\bm{\hat{y}}|\bm{\hat{z}}}(\bm{\hat{y}}|\bm{\hat{z}})\approx |\bm{\hat{y}}-\bm{\mu}|
\end{split}
\end{equation}

From the above deduction, the scales $\bm{\sigma}$ could be considered as the prediction of the absolute residual between the latent and means.

\subsection{Preliminary Improvements for Failed Predictions}
As discussed above, we can determine whether the predicted mean $\mu_i$ for $\hat{y}_i$ is accurate by $\sigma_i$: $\mu_i$ and $\hat{y}_i$ are more likely to have a noticeable difference if $\sigma_i$ is relatively large, which usually needs more bits for arithmetic encoding. The integration intervals for calculating likelihoods are typically at the ``tail'' of the entropy model for these ``failed predictions'', thus we can improve the entropy model:
\begin{itemize}
\item[$\bullet$]Adopting a more ``heavy-tailed'' entropy model, such as generalized Gaussian distribution with shape parameter $\beta<1$.
\item[$\bullet$]Sacrificing the likelihood interval of integers at the central for the likelihood interval of integers at the tail.
\end{itemize}

Based on the above discussion, we make preliminary improvements to SparsePCAC, which is a classical variational compression model for point cloud attribute compression with hyperprior and autoregressive context. As SparsePCAC uses Laplacian distribution as its entropy model, we first roughly assume that for $i$ with $\sigma_i>2$, the predicted mean $\mu_i$ for $\hat{y}_i$ is inaccurate, then use more heavy-tailed distributions and likelihood intervals closer to the center for these $i$ without retraining. 

Specifically, we directly use the trained model to obtain the latent and parameters of entropy models, but modify the likelihood calculation during encoding and decoding: for $0<\sigma_i \leq 2$, let
\begin{equation}
    p_{\hat{y}_i|\bm{\hat{z}}}(\hat{y}_i|\bm{\hat{z}})=\left[ \mathcal{L}(\mu_i, \sigma_i) * \mathcal{U}\left(-\frac{1}{2}, \frac{1}{2}\right) \right] (\hat{y}_i).
\end{equation}
For $2<\sigma_i$ and $|\hat{y}_i-\mu_i|\leq\frac{1}{2}$, let
\begin{equation}
    p_{\hat{y}_i|\bm{\hat{z}}}(\hat{y}_i|\bm{\hat{z}})=\left[ \mathcal{G}(\mu_i, \sigma_i,0.5) * \mathcal{U}\left(-\frac{1}{4}, \frac{1}{4}\right) \right] (\hat{y}_i).
    \label{formula:off1}
\end{equation}
For $2<\sigma_i$ and $|\hat{y}_i-\mu_i|>\frac{1}{2}$, let
\begin{equation}
\begin{split}
    p_{\hat{y}_i|\bm{\hat{z}}}(\hat{y}_i|\bm{\hat{z}})=
    \left[ \mathcal{G}(\mu_i, \sigma_i,0.5) * \mathcal{U}\left(-\frac{1}{2}, \frac{1}{2}\right) \right] (\hat{y}_i+\Delta)
    \label{formula:off2}
\end{split}
\end{equation}
where the offset $\Delta$ is:
\begin{equation}
    \Delta=\frac{1}{4}\operatorname{sgn}(\mu_i-\hat{y}_i).
\end{equation}
% \begin{equation}
% \begin{split}
%     p_{\hat{y}_i|\bm{\hat{z}}}(\hat{y}_i|\bm{\hat{z}})= \\
%     \left[ \mathcal{G}(\mu_i, \sigma_i,0.5) * \mathcal{U}\left(-\frac{1}{2}+\frac{1}{4}\operatorname{sgn}(\mu_i-\hat{y}_i), \frac{1}{2}+\frac{1}{4}\operatorname{sgn}(\mu_i-\hat{y}_i)\right) \right] (\hat{y}_i).
% \end{split}
% \end{equation} %无offset

% \begin{figure*}
% \begin{equation}
% % \begin{split}
%     p_{\hat{y}_i|\bm{\hat{z}}}(\hat{y}_i|\bm{\hat{z}})= \\
%     \begin{cases} 
%     \left[ \mathcal{L}(\mu_i, \sigma_i) * \mathcal{U}\left(-\frac{1}{2}, \frac{1}{2}\right) \right] (\hat{y}_i) 
%     & \text{if } 0<\sigma_i \leq 2, \\
%     \left[ \mathcal{G}(\mu_i, \sigma_i,0.5) * \mathcal{U}\left(-\frac{1}{4}, \frac{1}{4}\right) \right] (\hat{y}_i)
%     & \text{if }  2<\sigma_i \text{ and } |\hat{y}_i-\mu_i|\leq\frac{1}{2},\\
%    \left[ \mathcal{G}(\mu_i, \sigma_i,0.5) * \mathcal{U}\left(-\frac{1}{2}+\frac{1}{4}\operatorname{sgn}(\mu_i-\hat{y}_i), \frac{1}{2}+\frac{1}{4}\operatorname{sgn}(\mu_i-\hat{y}_i)\right) \right] (\hat{y}_i)
%    & \text{if }  2<\sigma_i \text{ and } |\hat{y}_i-\mu_i|>\frac{1}{2}.
%     \end{cases}
% % \end{split}
% \end{equation}
% \end{figure*} %双栏公式

We use generalized Gaussian distribution with $\beta=0.5$ for $i$ satisfying $\sigma_i>2$ in Equation \ref{formula:off1} and \ref{formula:off2}, and narrow the likelihood interval of $\lfloor\mu_i\rceil$, making the likelihood intervals of the remaining integers closer to the center.

\begin{figure}[ht]
\centering
\includegraphics[width=\columnwidth]{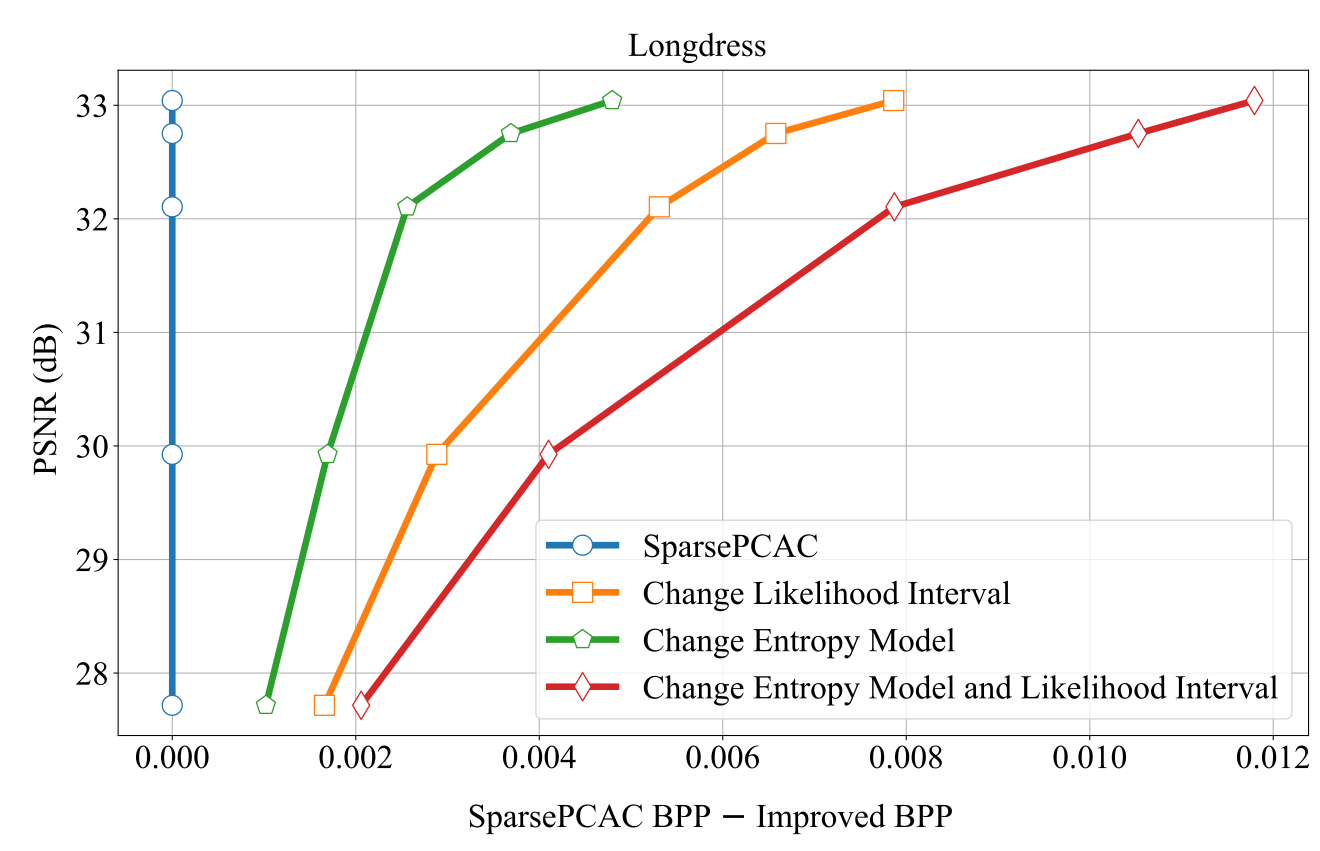}
% \captionsetup{font={small,bf,stretch=1.25}, justification=raggedright}
 \captionsetup{font={small}, singlelinecheck=off}
\caption{Results of preliminary experiment. To highlight the differences between different methods, the horizontal axis is calculated by subtracting improved bpp from the bpp of SparsePCAC.}
\label{fig:pre}
\end{figure}

Figure \ref{fig:pre} shows the results of the above improvements made to the SparsePCAC. From Figure \ref{fig:pre}, it can be seen that although the model structure has not been modified or retrained, there is still a noticeable improvement in encoding performance.

\section{Proposed Method}
\label{Proposed Method}

\begin{figure*}
\centering
\includegraphics[width=\textwidth]{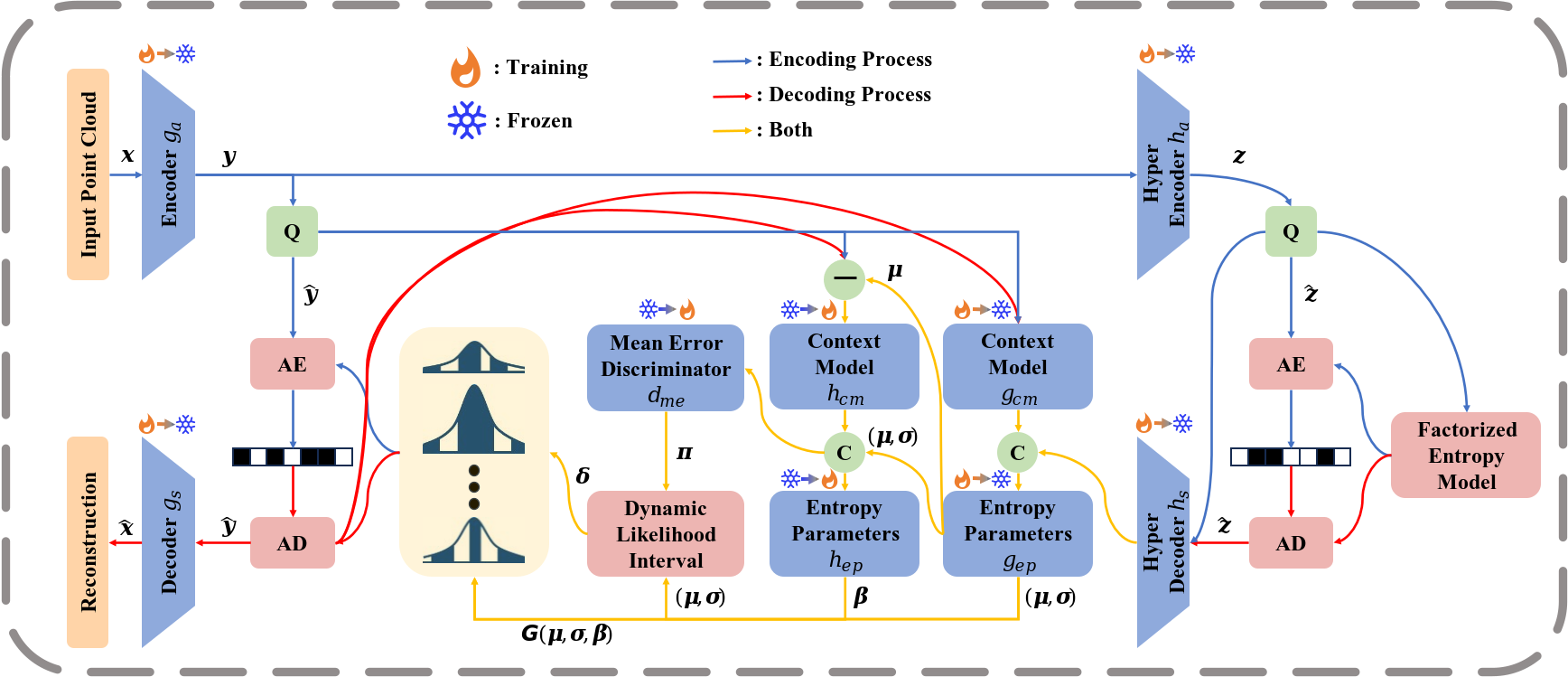}
% \captionsetup{font={small,bf,stretch=1.25}, justification=raggedright}
\captionsetup{font={small}}
\caption{Overall framework. We propose using the generalized Gaussian distribution as the entropy model to better estimate the probability of latents, and dynamically adjust likelihood intervals based on predicted entropy parameters. The entire network is trained by proposed two-step training strategy. \textbf{Q}  represents quantization and \textbf{C} represents concatenation. Blue and red lines represent the encoding and decoding steps respectively, while yellow lines represent the steps that need to be performed in both encoding and decoding.}
\label{fig:framework}
\end{figure*}

The overall framework of the proposed method is illustrated in Figure \ref{fig:framework} based on the above deduction and experimental results. We utilize the generalized Gaussian entropy model and dynamic likelihood interval to encode latents more efficiently, which will be detailed discussed next.

\subsection{Generalized Gaussian Entropy Model}
For the input point cloud $\bm{x}$, we can first obtain quantized latent $\bm{\hat{y}}$ and hyper latent $\bm{\hat{z}}$ through autoencoders $(g_a,g_s)$ and $(h_a,h_s)$:
\begin{equation}
\begin{split}
    &\bm{\hat{y}}=Q(\bm{y})=Q(g_a(\bm{x})) \\
    &\bm{\hat{z}}=Q(\bm{z})=Q(h_a(\bm{y})),
\end{split}
\end{equation}
then use the generalized Gaussian distribution as the entropy model for $\bm{\hat{y}}$:
\begin{equation}
    p_{\bm{\hat{y}}|\bm{\hat{z}}}(\bm{\hat{y}}|\bm{\hat{z}})\sim\mathcal{G}(\bm{\mu},\bm{\sigma},\bm{\beta}).
\end{equation}

We estimate the mean and scale of the generalized Gaussian entropy model by combining hyperprior and context , which is similar to SparsePCAC:
\begin{equation}
    (\mu_i, \sigma_i) = g_{ep}( g_{cm}(\bm{\hat{y}}_{<i}),h_s(\bm{\hat{z}})).
\end{equation}
As the likelihood is determined by the probability density function in Equation \ref{pdf} which depends on $\bm{x}-\bm{\mu}$ (i.e., $\bm{\hat{y}}-\bm{\mu}$), $\bm{\sigma}$ and $\bm{\beta}$, the optimal $\bm{\beta}$ can be determined given $\bm{\hat{y}}-\bm{\mu}$ and $\bm{\sigma}$ in order to maximize the likelihood. Different from $\mu_i$ and $\sigma_i$, $\hat{y}_i$ is unknown when determining $\beta_i$, so we propose another context model $h_{cm}$ to extract context from the absolute difference between predicted $\bm{\mu}$ and decoded $\bm{\hat{y}}$:
\begin{equation}
    \mathcal{C}=h_{cm}(|\bm{\hat{y}}-\bm{\mu}|_{<i}).
\end{equation}
Therefore, we combine $(\bm{\mu},\bm{\sigma})$ and context $\mathcal{C}$ to estimate $\bm{\beta}$:
\begin{equation}
\begin{split}
    \beta_i 
    &= h_{ep}(\mathcal{C},\bm{\mu}_{\leq i},\bm{\sigma}_{\leq i}) \\
    &= h_{ep}(h_{cm}(|\bm{\hat{y}}-\bm{\mu}|_{<i}),\bm{\mu}_{\leq i},\bm{\sigma}_{\leq i}).
\end{split}
\end{equation}

\begin{table*}
\centering
\scalebox{0.75}{% 缩小到原来的80%
\scriptsize
\label{tab:my-table}
\begin{tabular}{c|c|c|c|c|c|c|c}
\textbf{Encoder} $g_\text{a}$* & \textbf{Decoder} $g_\text{s}$* & \textbf{Hyper Encoder} $h_\text{a}$* & \textbf{Hyper Decoder} $h_\text{s}$* & \textbf{Context Prediction} $g_\text{cm}$*, $h_\text{cm}$ & \textbf{Entropy Parameters} $g_\text{ep}$* & \textbf{Entropy Parameters} $h_\text{ep}$& \textbf{Mean Error Discriminator} $d_\text{me}$\\
\midrule
SConv $\text{3}^\text{3}$ c64 & 
TSConv $\text{3}^\text{3}$ c128 s2$\uparrow$ & 
SConv $\text{3}^\text{3}$ c128 & 
TSConv $\text{3}^\text{3}$ c128 s2$\uparrow$ & 
Masked $\text{5}^\text{3}$ c256 & 
SConv $\text{1}^\text{3}$ c384&
SConv $\text{1}^\text{3}$ c384& 
SConv $\text{1}^\text{3}$ c384\\

SConv $\text{3}^\text{3}$ c128 s2$\downarrow$ & 
SConv $\text{3}^\text{3}$ c128 & 
Leaky ReLU & 
SConv $\text{3}^\text{3}$ c128 &  
& 
Leaky ReLU &
Leaky ReLU&
Leaky ReLU\\

Leaky ReLU &
Leaky ReLU &
SConv $\text{3}^\text{3}$ c128&
Leaky ReLU &
&
SConv $\text{1}^\text{3}$ c256 &
SConv $\text{1}^\text{3}$ c256&
SConv $\text{1}^\text{3}$ c256\\

SConv $\text{3}^\text{3}$ c128 &
TSConv $\text{3}^\text{3}$ c128 s2$\uparrow$ &
SConv $\text{3}^\text{3}$ c128 s2$\downarrow$ &
TSConv $\text{3}^\text{3}$ c256 s2$\uparrow$ &
&
Leaky ReLU &
Leaky ReLU&
Leaky ReLU\\

SConv $\text{3}^\text{3}$ c128 s2$\downarrow$ &
SConv $\text{3}^\text{3}$ c128 &
Leaky ReLU &
SConv $\text{3}^\text{3}$ c256 &
&
SConv $\text{1}^\text{3}$ c256 &
SConv $\text{1}^\text{3}$ c128&
SConv $\text{1}^\text{3}$ c128\\

Leaky ReLU &
Leaky ReLU &
SConv $\text{3}^\text{3}$ c128&
Leaky ReLU&
& 
&
&
Leaky ReLU\\

SConv $\text{3}^\text{3}$ c128 &
TSConv $\text{3}^\text{3}$ c64 s2$\uparrow$ &
SConv $\text{3}^\text{3}$ c128 s2$\downarrow$ & 
SConv $\text{3}^\text{3}$ c256& 
& 
& 
&
Sigmoid\\

SConv $\text{3}^\text{3}$ c128 s2$\downarrow$ &
SConv $\text{3}^\text{3}$ c3 &
& 
& 
& 
&
&\\

\end{tabular}}
\captionsetup{font={small},justification=raggedright, singlelinecheck=off}
    \caption{Network Architecture. The modules marked with * refer to SparsePCAC and will be replaced with corresponding modules in subsequent experiments while $h_{cm}$ we propose is consistent with $g_{cm}$. ``SConv'' is sparse convolution and ``TSConv'' is transposed sparse convolution.}

\label{tab:arch}
\end{table*}

\subsection{Dynamic Likelihood Interval}
\label{Dynamic Likelihood Interval}
In current methods, when performing arithmetic encoding on $\hat{y}_i$ using a continuous entropy model $\mathcal{G}(\mu_i, \sigma_i,\beta_i)$, the probability of integer $n$ is determined based on fix likelihood intervals:
\begin{equation}
\begin{split}
     P(\hat{y}_i=n)&=\left[ \mathcal{G}(\mu_i, \sigma_i,\beta_i) * \mathcal{U}\left(-\frac{1}{2}, \frac{1}{2}\right) \right](n) \\
     &=\int_{n-\frac{1}{2}}^{n+\frac{1}{2}}  \mathcal{G}(y|\mu_i, \sigma_i,\beta_i) \, dy.
\end{split}
\end{equation}
Results of preliminary experiment indicate that a fixed likelihood interval does not fully utilize the effective information contained in the parameters of the entropy model. In fact, we can determine  whether the prediction for $\hat{y}_i$ is accurate based on the entropy model parameters and decoded latent. If the prediction is inaccurate, indicated by a large $|\hat{y}_i-\mu_i|$, we can obtain a more precise discrete probability distribution for arithmetic encoding by shortening the likelihood interval for the central region integers, which causes bigger likelihoods of the integers in the marginal regions as the likelihood intervals of these integers converge toward the center. Conversely, if the prediction is accurate, we can extend the likelihood interval for the central region integers by sacrificing the likelihood interval of the marginal region integers.

However, it is challenging to accurately determine whether the likelihood interval of each integer should be extended or shortened. Therefore, we consider scaling the likelihood interval of $\lfloor\mu_i\rceil$ while translating the likelihood intervals of the remaining integers to obtain a more accurate discrete probability distribution for the sake of model simplicity and effectiveness, which is illustrated in Figure \ref{fig:dli}.

As discussed above, we should first determine the accuracy of the prediction $\mu_i$ for $\hat{y}_i$. Therefore, we estimate the probability of $\hat{y}_i=\lfloor\mu_i\rceil$ using Mean Error Discriminator $d_{me}$ based on $(\bm{\mu},\bm{\sigma})$ and context $\mathcal{C}$:
\begin{equation}
\begin{split}
    \pi_i
    :&=P(\hat{y}_i=\lfloor\mu_i\rceil) \\
    &= d_{me}(\mathcal{C},\bm{\mu}_{\leq i},\bm{\sigma}_{\leq i}) \\
    &= d_{me}(h_{cm}(|\bm{\hat{y}}-\bm{\mu}|_{<i}),\bm{\mu}_{\leq i},\bm{\sigma}_{\leq i}).
\end{split}
\end{equation}

Next, We determine the likelihood intervals of $\lfloor\mu_i\rceil$ and the remaining integers separately. In order to determine the likelihood interval of $\lfloor\mu_i\rceil$:
\begin{equation}
    LI(\lfloor\mu_i\rceil|\mu_i,\sigma_i,\pi_i)=\mathcal{U}\left(-\frac{\delta_i}{2}, \frac{\delta_i}{2}\right) (\lfloor\mu_i\rceil),
\end{equation}
we set the range for $\delta_i$ as follow:
\begin{equation}
   \frac{1}{1+\sigma_i}\leq \delta_i\leq \frac{1}{1-e^{-\sigma_i}}.
   \label{range}
\end{equation}
Equation \ref{range} uses $\sigma$ to determine the upper bound of stretching ($\frac{1}{1-e^{-\sigma_i}}$, with a range of $(1, +\infty)$) and the lower bound of squeezing ($\frac{1}{1+\sigma_i}$,with a range of $(0, 1)$) of the scaling ratio $\delta_i$. Since larger $\sigma_i$ means inaccurate $\mu_i$, it is better to reduce the likelihood interval of $\mu_i$ at this time, so Equation \ref{range} ensures that both upper bound and lower bound will decrease when $\sigma_i$ increases. By combining the bounds of $\delta_i$ and the probability $\pi_i$, the final $\delta_i$ is:
\begin{equation}
    \delta_i = 
\begin{cases} 
1+(2\pi_i-1)(\frac{1}{1-e^{-\sigma_i}}-1) & \text{if } \pi_i\geq \frac{1}{2}\\ 
1-(1-2\pi_i)(1-\frac{1}{1+\sigma_i}) & \text{if } \pi_i<\frac{1}{2} 
\end{cases}.
\label{fomula:delta}
\end{equation}
Equation \ref{fomula:delta} uses the relationship between $\pi_i$ and $0.5$ to determine whether to stretch or squeeze, and uses $|2\pi_i-1|$ (to make the whole function continuous) and $\sigma_i$ to determine the final $\delta_i$. As shown in Figure \ref{delta}, for $\pi_i\geq \frac{1}{2}$, we consider that $\hat{y}_i$ is more likely to be in the tail of the entropy model, which leads us to shorten the likelihood interval of $\lfloor\mu_i\rceil$. Conversely, we expand the likelihood interval of $\lfloor\mu_i\rceil$ for $\pi_i<\frac{1}{2}$.

Naturally, for integer $n\neq \lfloor\mu_i\rceil$, we set its likelihood interval as:
\begin{equation}
    LI(n|\mu_i,\sigma_i,\pi_i) = 
\begin{cases} 
\mathcal{U}\left(\frac{\delta_i}{2}-1, \frac{\delta_i}{2}\right) (n) & \text{if } n>\lfloor\mu_i\rceil\\ 
\mathcal{U}\left(-\frac{\delta_i}{2}, -\frac{\delta_i}{2}+1\right) (n) & \text{if } n<\lfloor\mu_i\rceil\\ 
\end{cases}.
\end{equation}

\subsection{Network Architecture}
The structures of different neural network modules in our method are outlined in Table \ref{tab:arch}. Please note that the modules marked with * are the same as SparsePCAC for example. In subsequent experiments, we will replace the structures of these modules with those of corresponding modules in different existing methods.

\subsection{Loss Function}
We still use the rate-distortion cost with MSE distortion as the loss function for training:
\begin{equation}
\begin{split}
    \mathcal{L}&=R + \lambda \cdot D \\
    &= E_{\bm{x} \sim p_{\bm{x}}} \left[ -\log_{2} p_{\bm{\hat{y}}|\bm{\hat{z}}}(\bm{\hat{y}}|\bm{\hat{z}}) - \log_{2} p_{\bm{\hat{z}}}(\bm{\hat{z}}) \right] \\
    &+ \lambda \cdot E_{\bm{x} \sim p_{\bm{x}}} [d(\bm{x}, \bm{\hat{x}})],
\end{split}
\end{equation}
but the likelihoods of latent $\bm{\hat{y}}$ need to be calculated according to generalized gaussian entropy model and dynamic likelihood intervals:
\begin{equation}
\begin{split}
    p_{\bm{\hat{y}}|\bm{\hat{z}}}(\bm{\hat{y}}|\bm{\hat{z}})=\prod_i \left[ \mathcal{G}(\mu_i, \sigma_i,\beta_i) * LI\left(\mu_i,\sigma_i,\pi_i\right) \right] (\hat{y}_i).
\end{split}
\end{equation}
\section{Experimental Results}

\begin{figure*}[htbp]
    \centering
    \begin{subfigure}[b]{0.3\textwidth}
        \includegraphics[width=\textwidth]{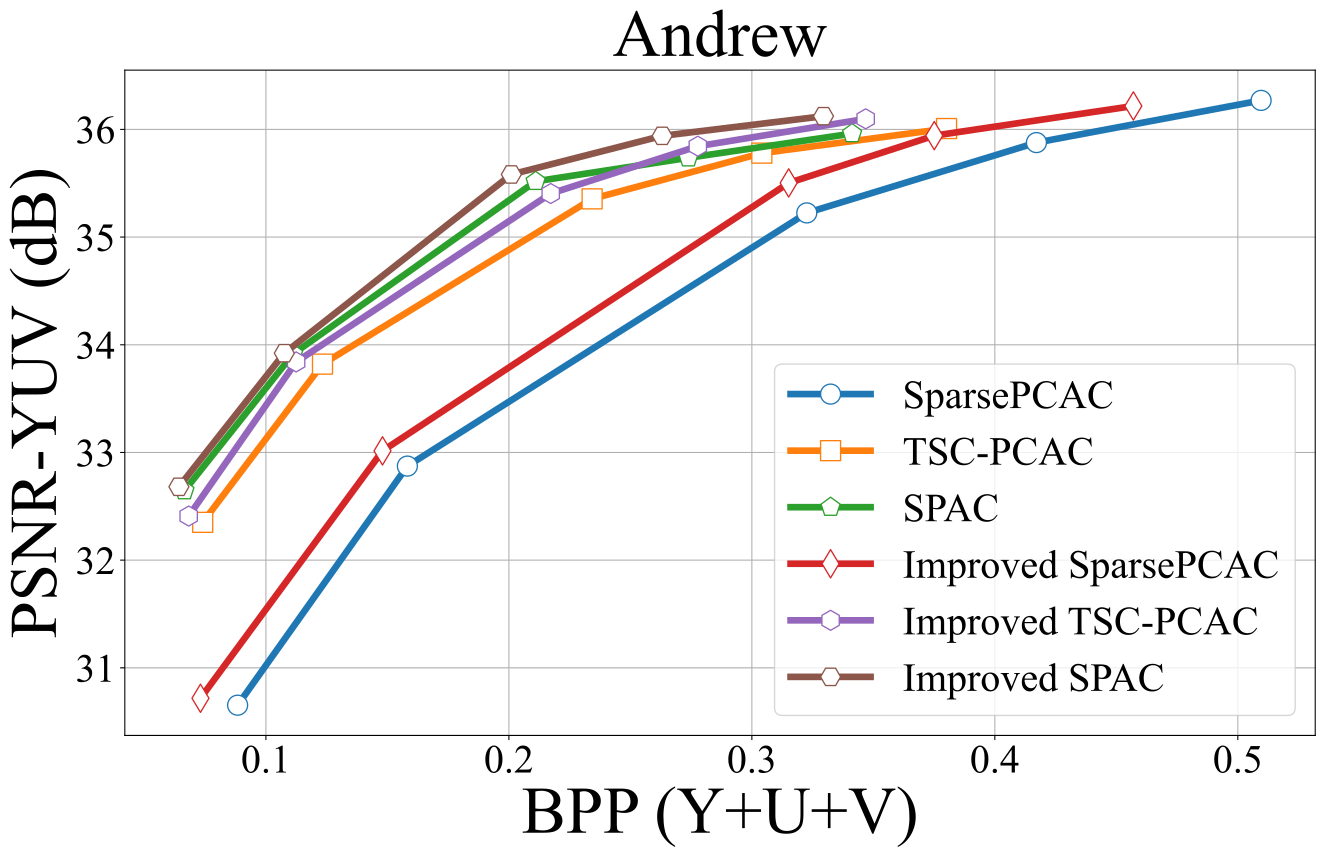}
    \end{subfigure}
    \hfill
    \begin{subfigure}[b]{0.3\textwidth}
        \includegraphics[width=\textwidth]{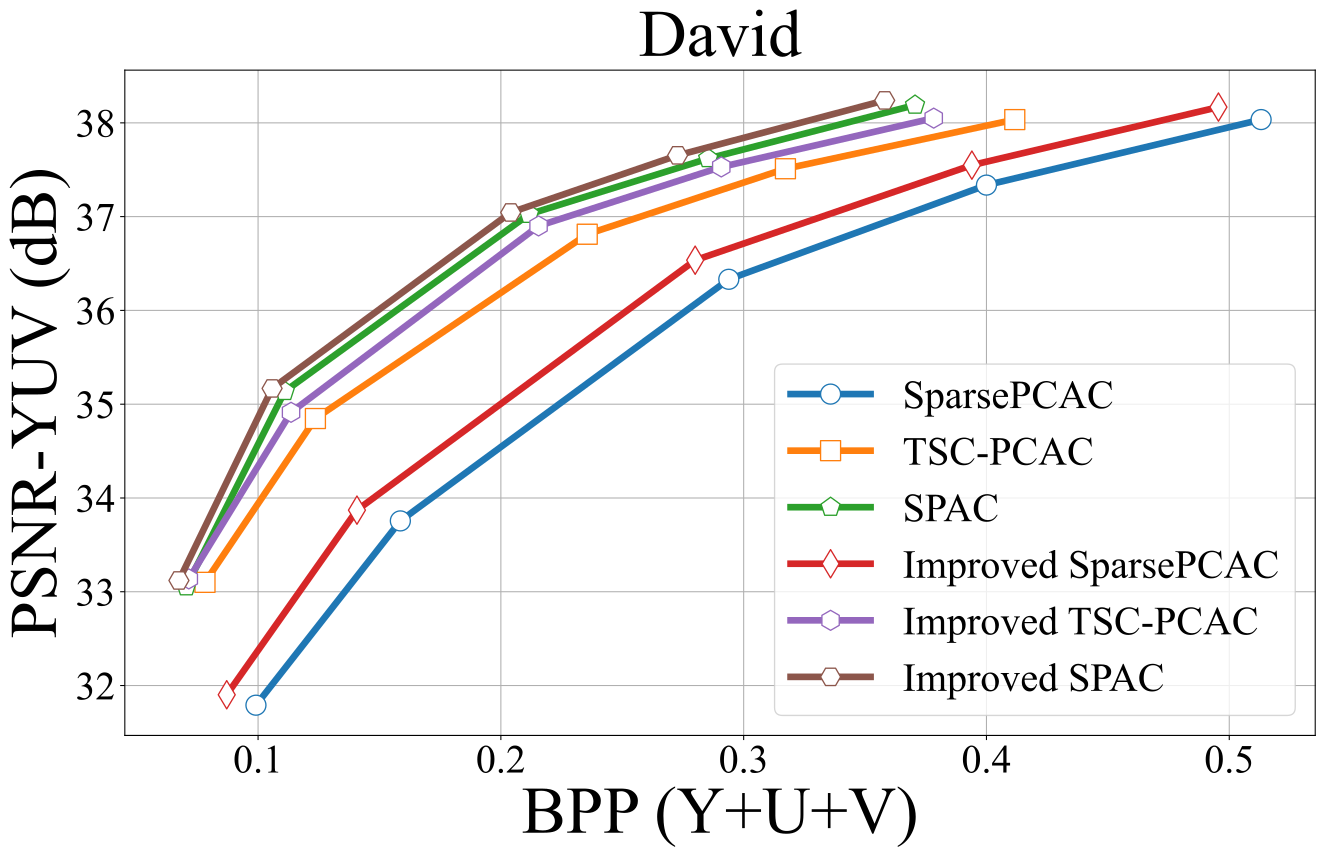}
    \end{subfigure}
    \hfill
    \begin{subfigure}[b]{0.3\textwidth}
        \includegraphics[width=\textwidth]{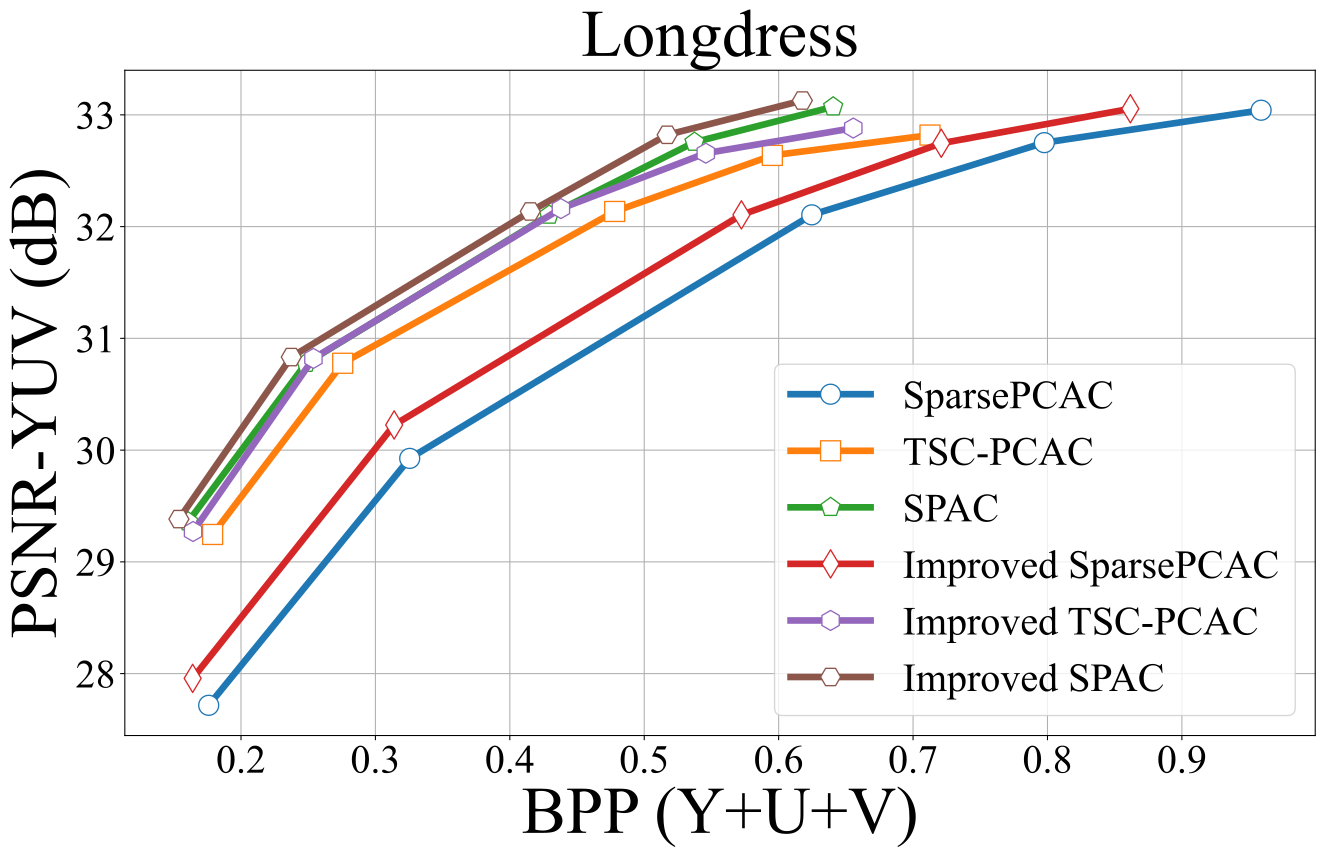}
    \end{subfigure}
    \hfill
    \begin{subfigure}[b]{0.3\textwidth}
        \includegraphics[width=\textwidth]{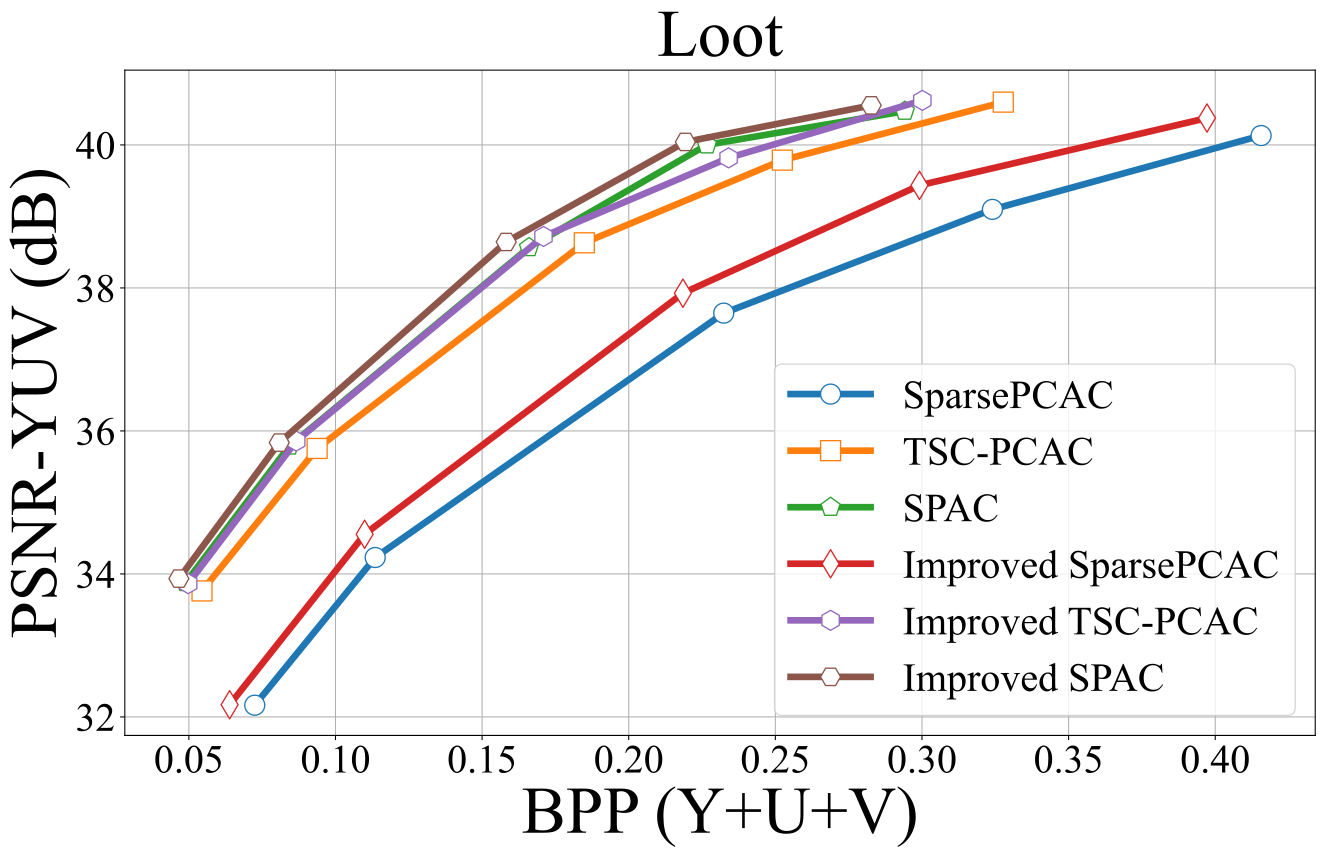}
    \end{subfigure}
    \hfill
    \begin{subfigure}[b]{0.3\textwidth}
        \includegraphics[width=\textwidth]{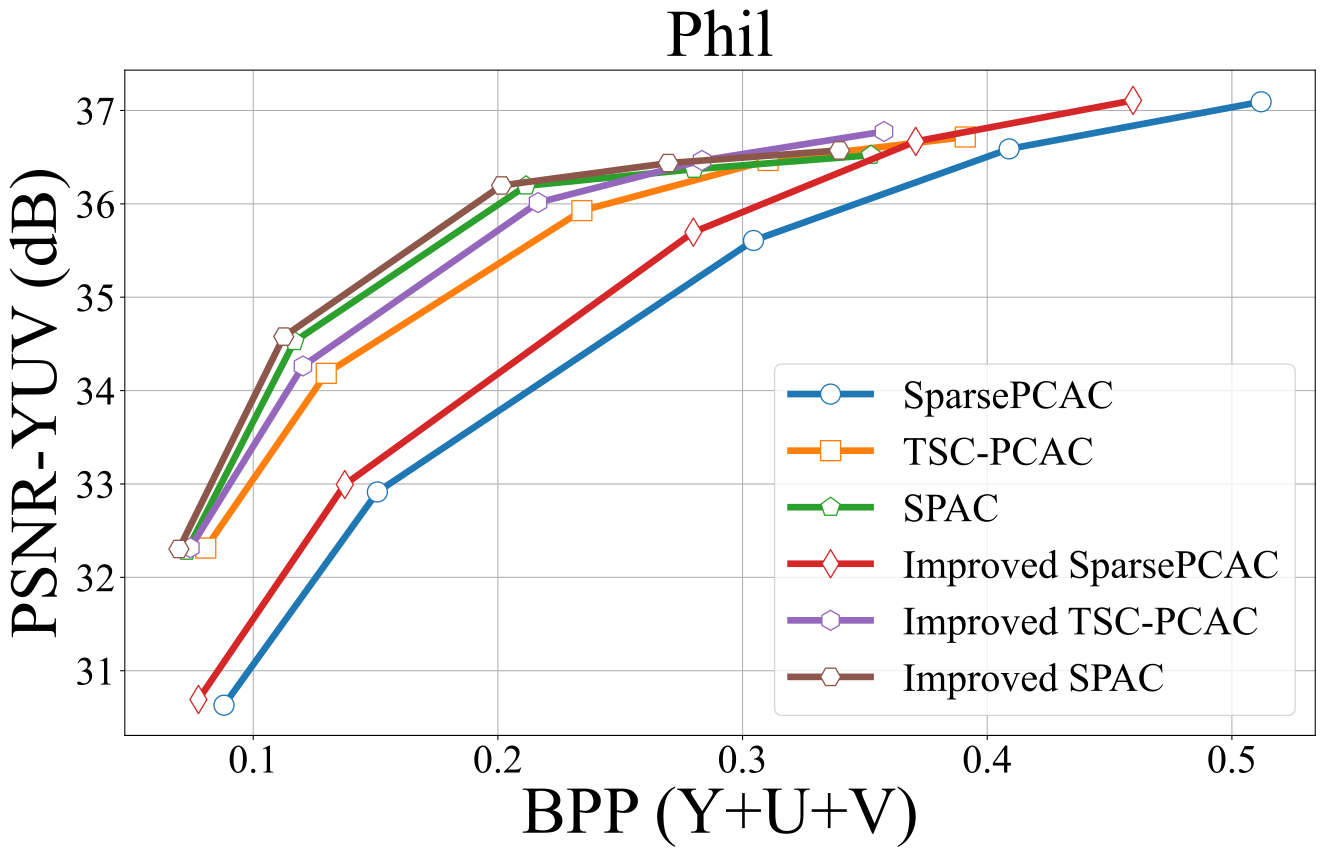}
    \end{subfigure}
 \hfill
       \begin{subfigure}[b]{0.3\textwidth}
       \includegraphics[width=\textwidth]{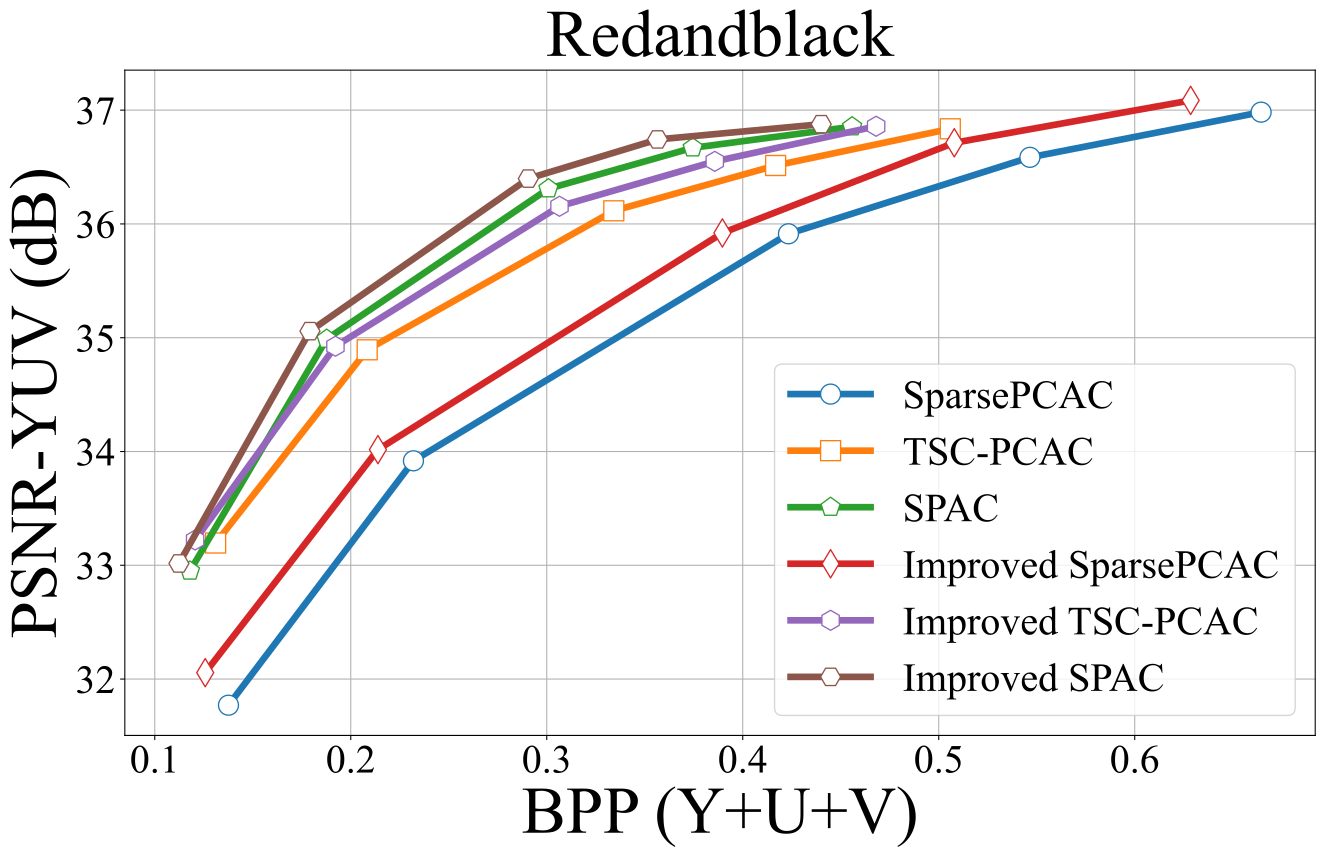}
    \end{subfigure}
 \hfill
\begin{subfigure}[b]{0.3\textwidth}
        \includegraphics[width=\textwidth]{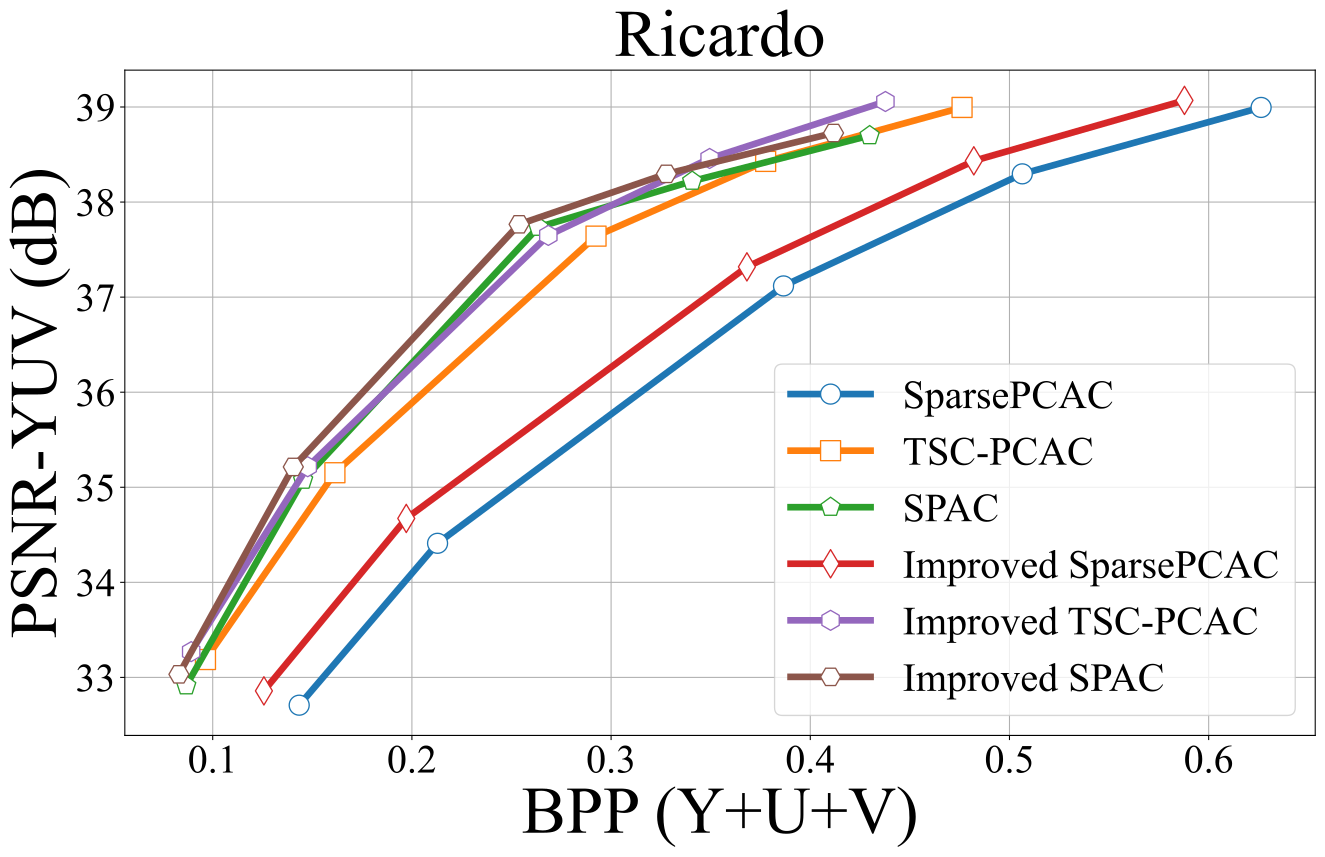}
    \end{subfigure}
    \hfill
    \begin{subfigure}[b]{0.3\textwidth}
        \includegraphics[width=\textwidth]{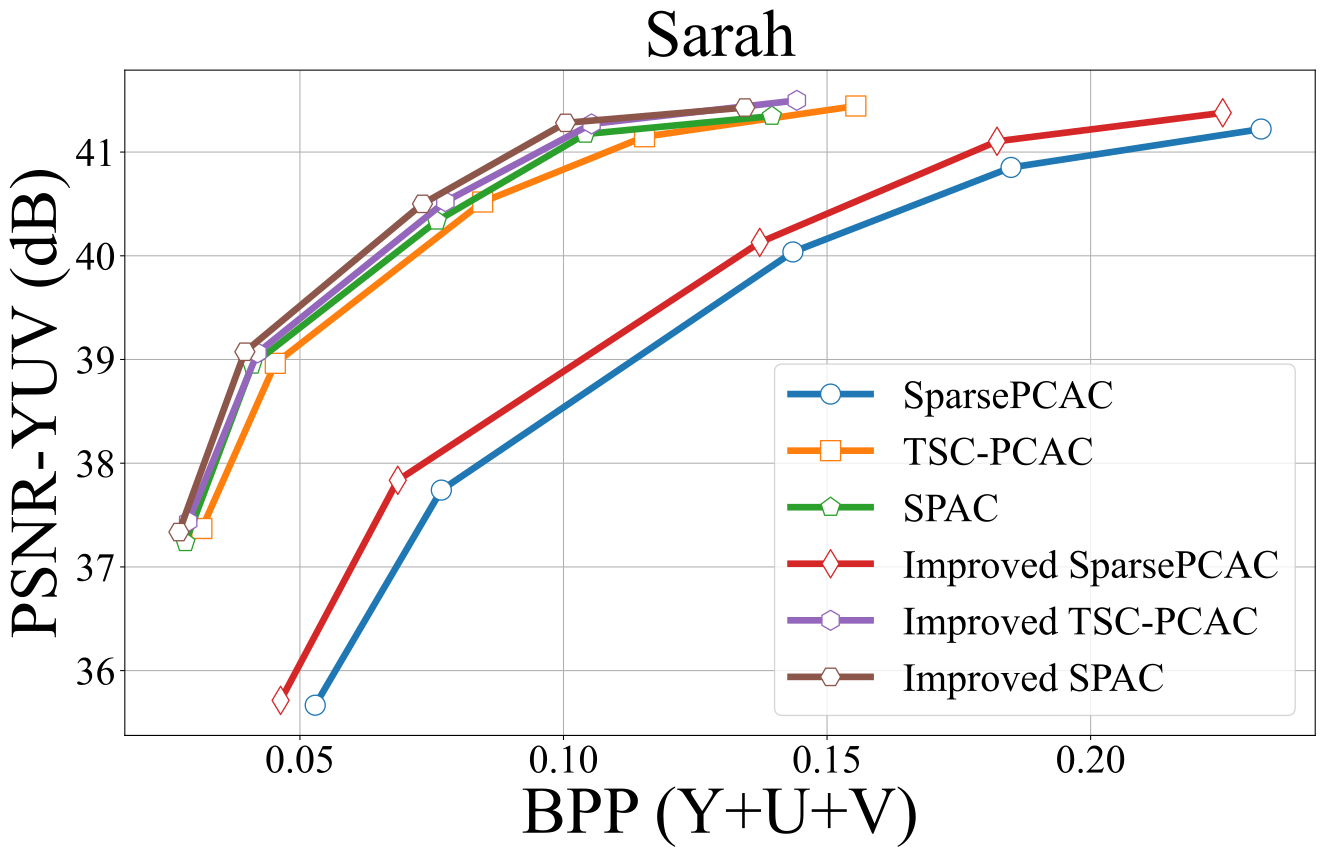}
    \end{subfigure}
 \hfill
       \begin{subfigure}[b]{0.3\textwidth}
       \includegraphics[width=\textwidth]{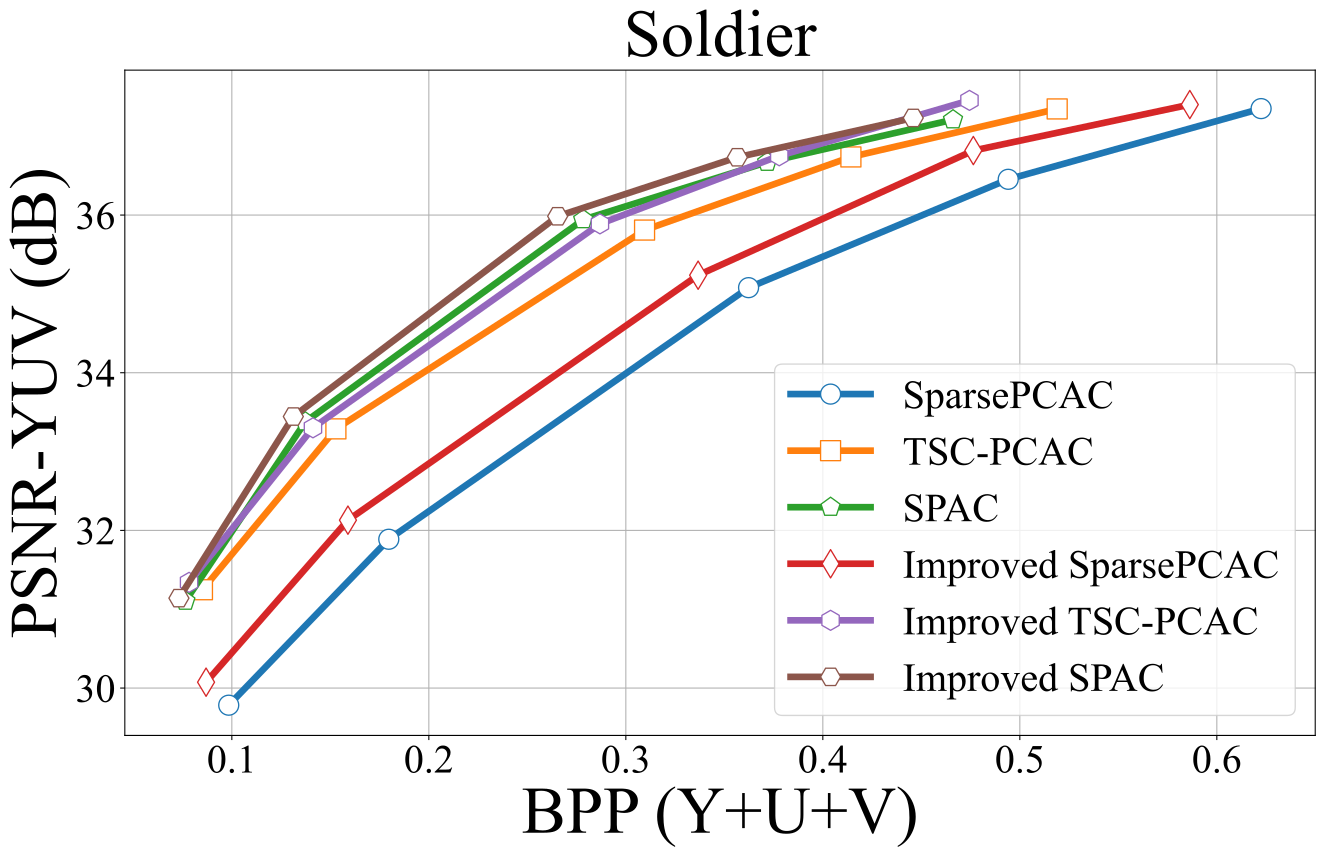}
    \end{subfigure}
    \captionsetup{font={small}, singlelinecheck=off}
    \caption{Demonstration of rate-distortion curves for the proposed method applied to three baseline models on nine point cloud sequences.}
    \label{fig:rd}
\end{figure*}

%%%%%%%%%%%%%%%%%%arXiv
\begin{figure*}[htbp]
\begin{minipage}{1\textwidth}
\centering
        \includegraphics[width=\textwidth]{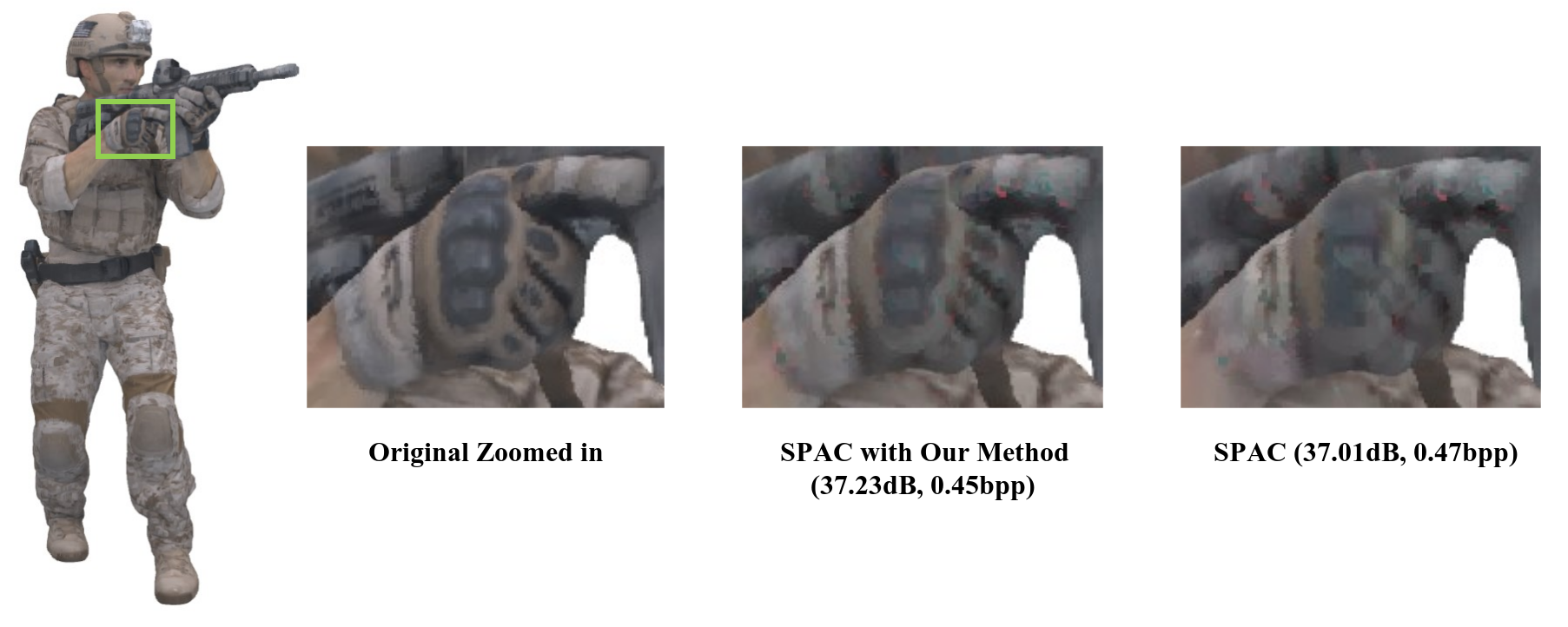}
        \subcaption{\textit{Soldier}}
    \end{minipage}
    \hfill
\begin{minipage}{1\textwidth}
\centering
        \includegraphics[width=\textwidth]{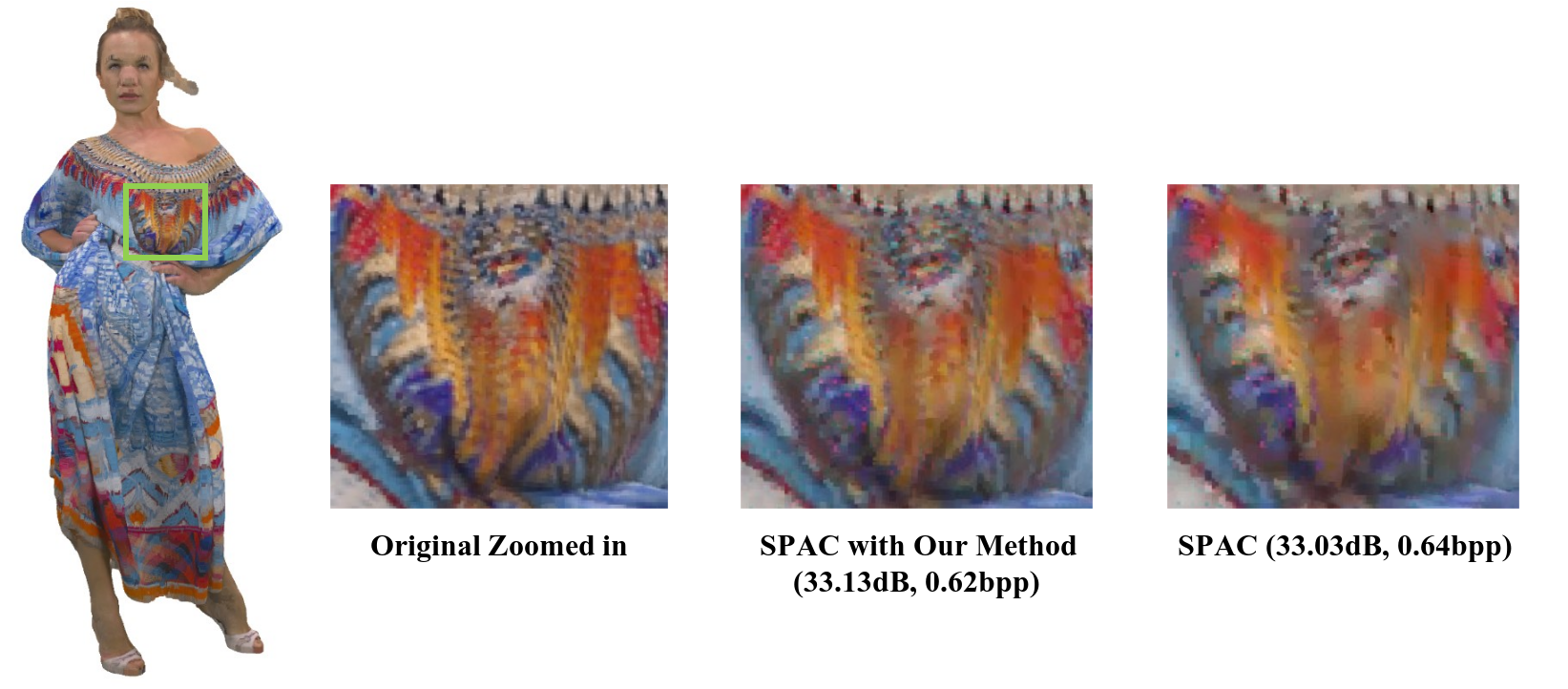}
        \subcaption{\textit{Longdress}}
    \end{minipage}
    \hfill
\begin{minipage}{1\textwidth}
\centering
        \includegraphics[width=\textwidth]{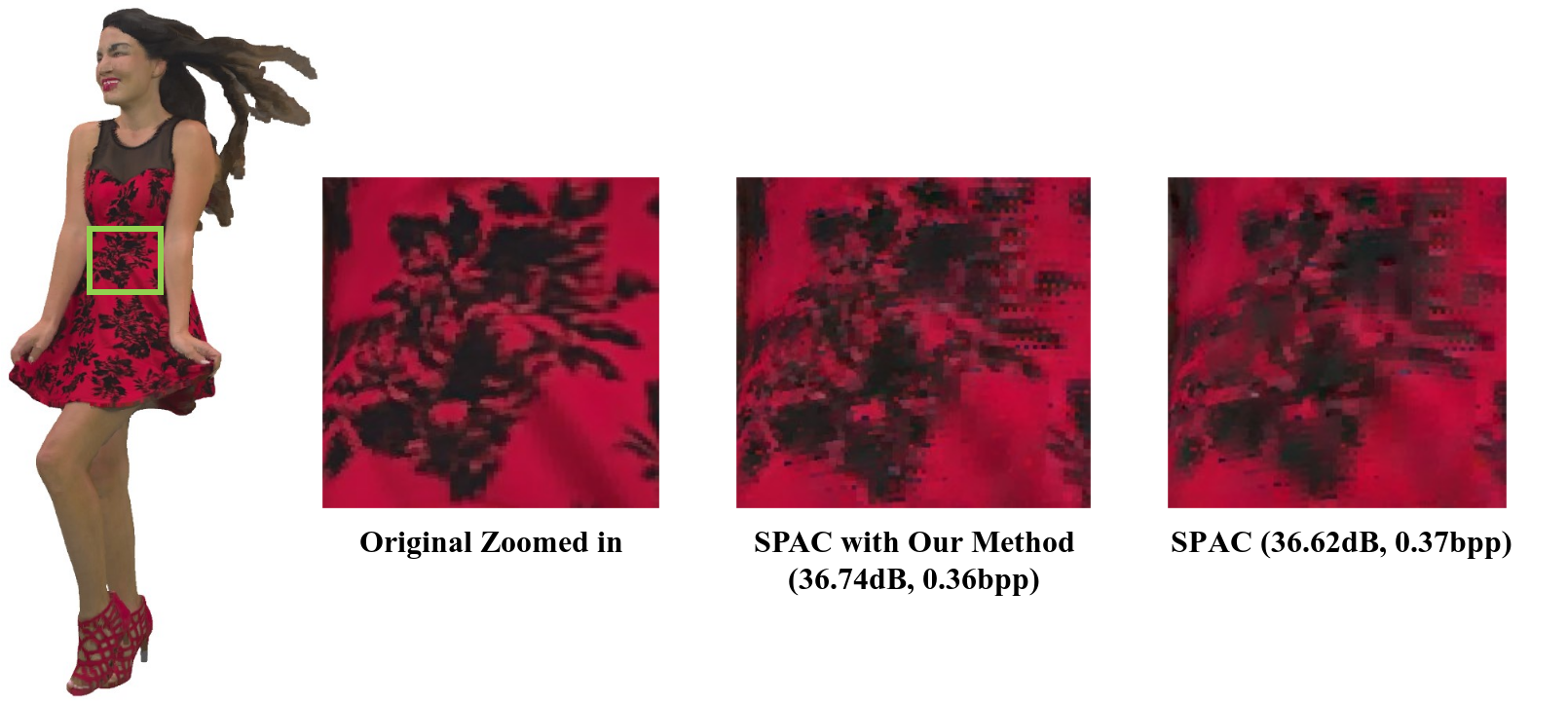}
        \subcaption{\textit{Redandblack}}
    \end{minipage}
\captionsetup{font={small}, singlelinecheck=off}
  \caption{Visual quality comparison.}
  \label{fig:visual}
\end{figure*}
%%%%%%%%%%%%%%%%%%%%%%%%%%%%arXiv
\subsection{Datastes}
To show the model's transferability, we construct the training and testing datasets as follow:

\textbf{Training Dataset.} We use ShapeNet \cite{chang2015shapenet} and COCO \cite{cocodataset} to construct the training dataset as in \cite{wang2022sparse}. We randomly sample points from the meshes in ShapeNet, apply random rotation to them, and then quantize the coordinates into 8-bit integers to represent the geometry of point clouds. The number of points in each point cloud ranges from 50,000 to 100,000. Next, we randomly select images from COCO and project them onto the point clouds obtained through the aforementioned steps as the attributes of these point clouds. We generate 12,000 samples for traning by this approach.

\begin{figure}[htbp]
    \centering
    \begin{subfigure}[b]{0.15\textwidth}
        \includegraphics[width=\textwidth]{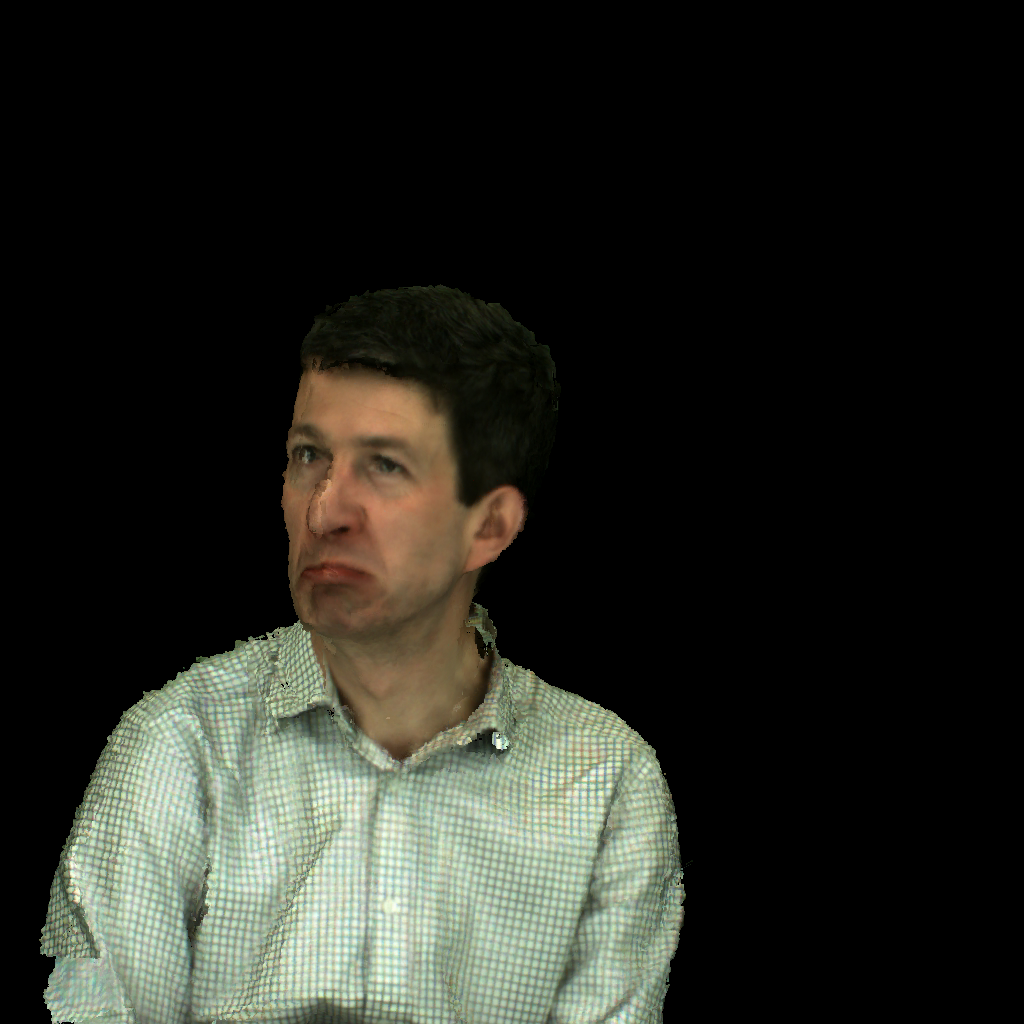}
        \caption{\textit{Andrew}}
    \end{subfigure}
    \hfill
    \begin{subfigure}[b]{0.15\textwidth}
        \includegraphics[width=\textwidth]{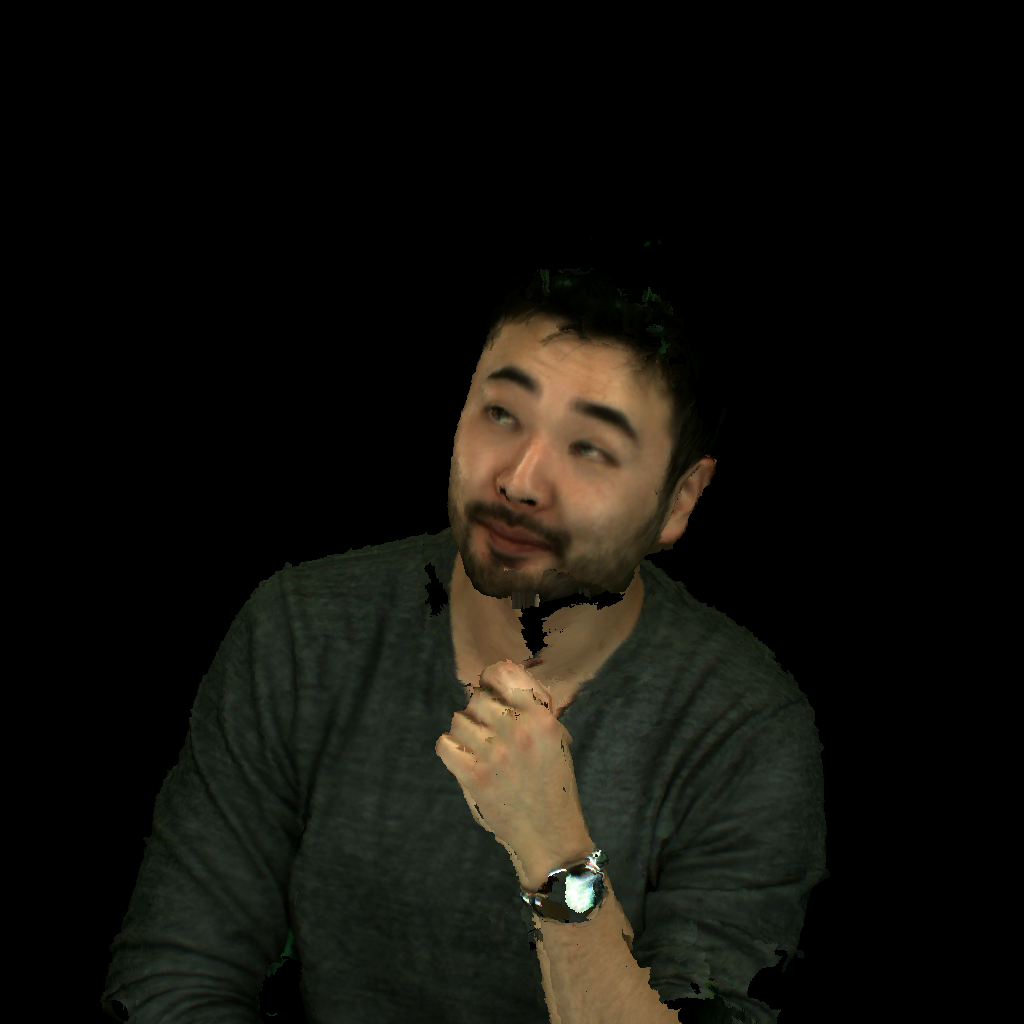}
        \caption{\textit{David}}
    \end{subfigure}
    \hfill
    \begin{subfigure}[b]{0.15\textwidth}
        \includegraphics[width=\textwidth]{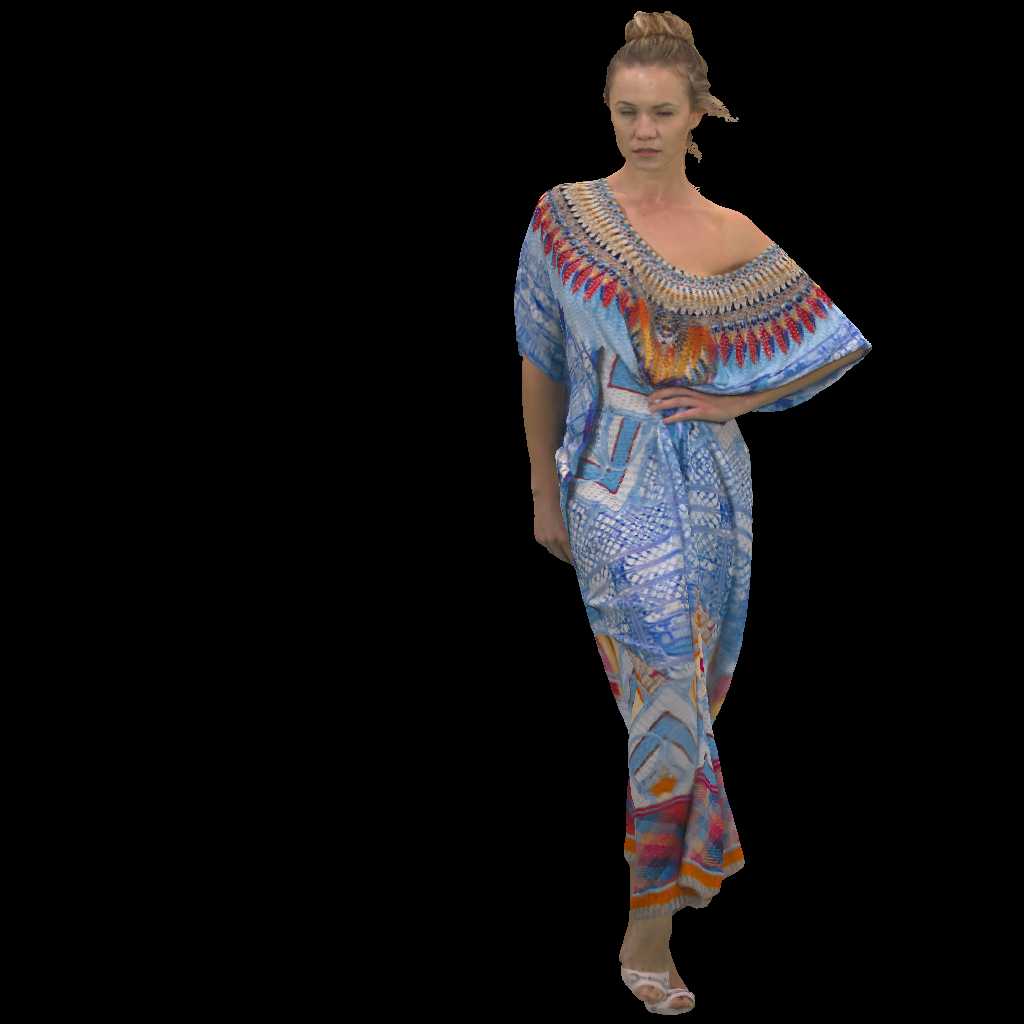}
        \caption{\textit{Longdress}}
    \end{subfigure}
    \hfill
    \begin{subfigure}[b]{0.15\textwidth}
        \includegraphics[width=\textwidth]{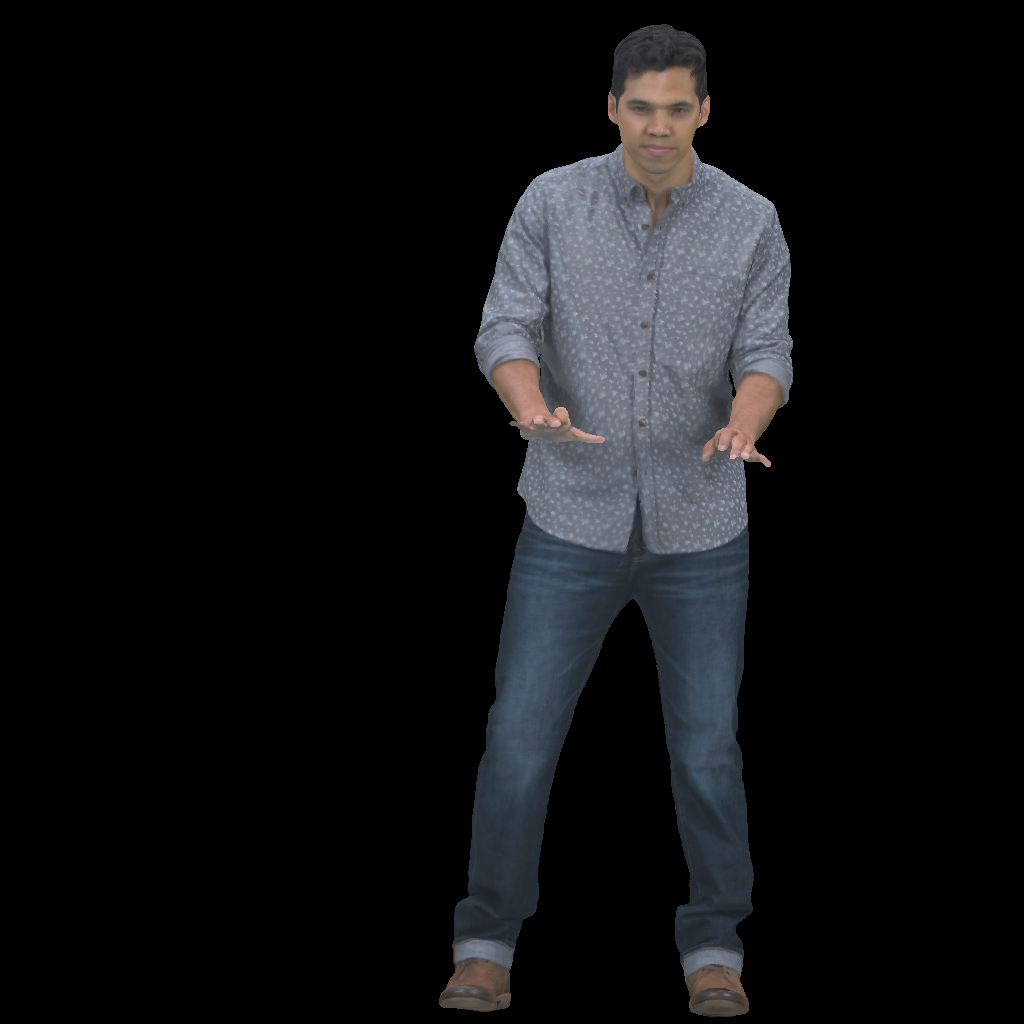}
        \caption{\textit{Loot}}
    \end{subfigure}
    \hfill
    \begin{subfigure}[b]{0.15\textwidth}
        \includegraphics[width=\textwidth]{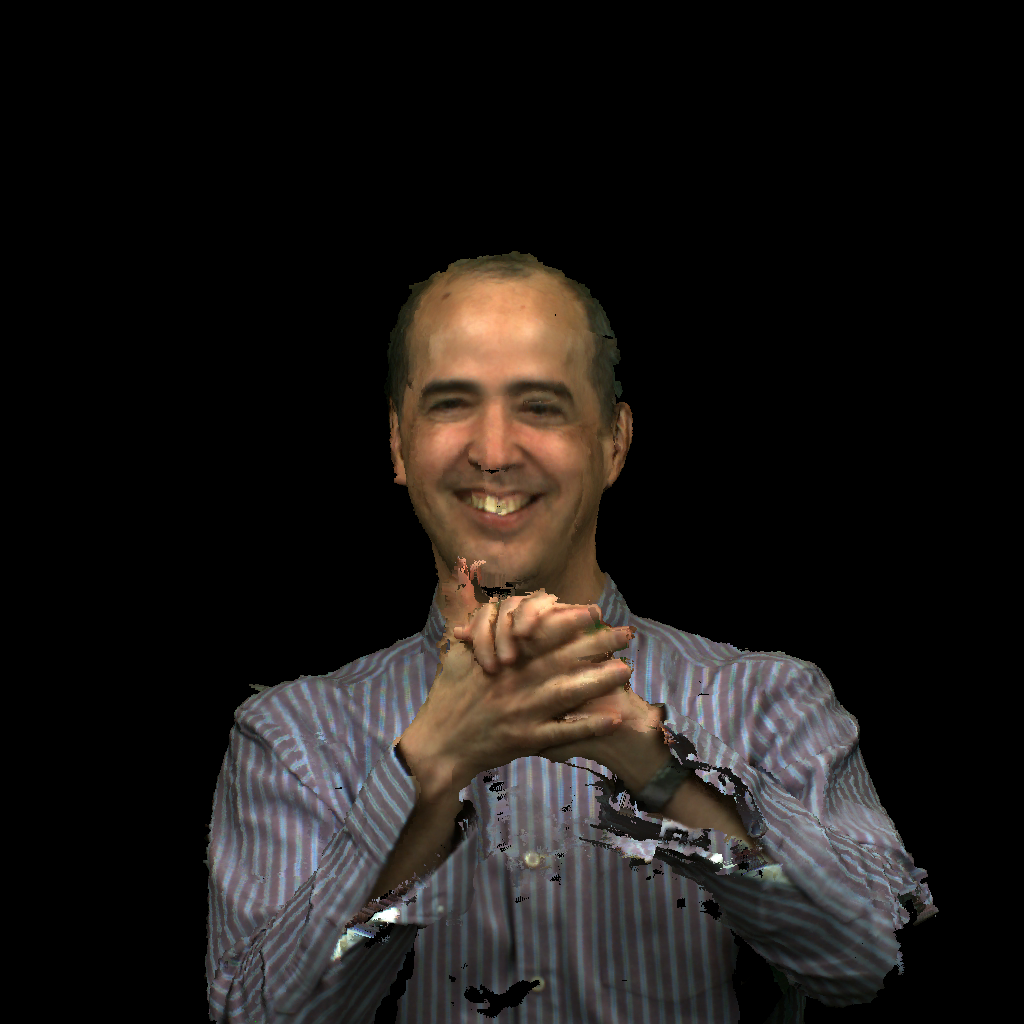}
        \caption{\textit{Phil}}
    \end{subfigure}
 \hfill
    \begin{subfigure}[b]{0.15\textwidth}
        \includegraphics[width=\textwidth]{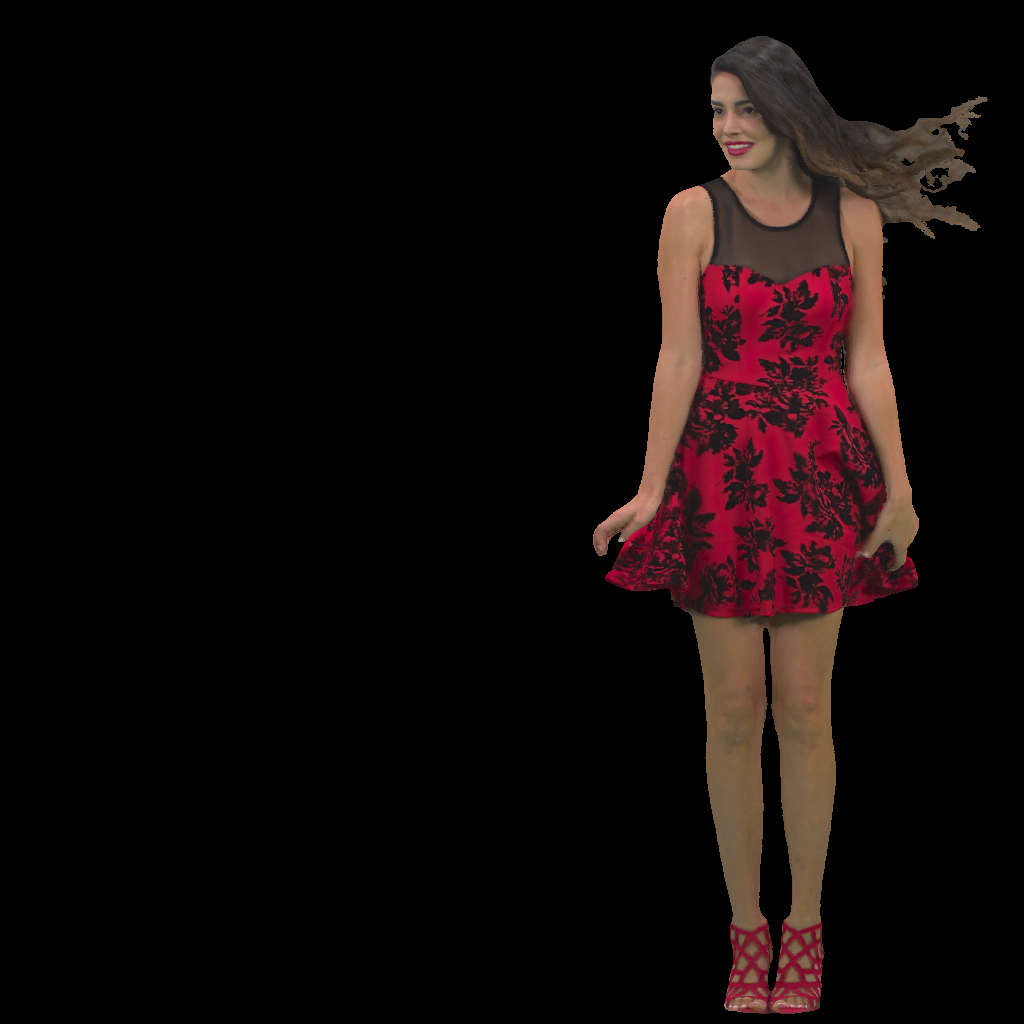}
        \caption{\textit{Redandblack}}
    \end{subfigure}
 \hfill
    \begin{subfigure}[b]{0.15\textwidth}
        \includegraphics[width=\textwidth]{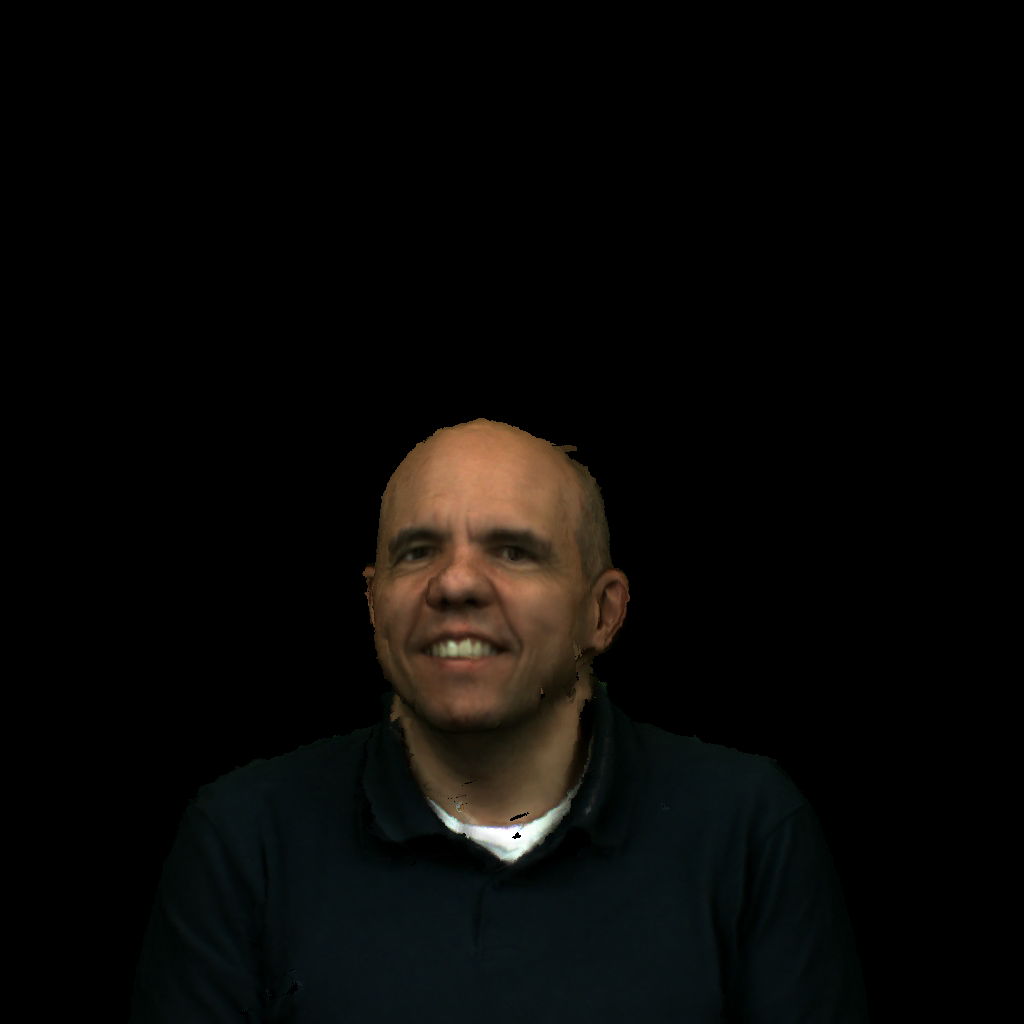}
        \caption{\textit{Ricardo}}
    \end{subfigure}
 \hfill
    \begin{subfigure}[b]{0.15\textwidth}
        \includegraphics[width=\textwidth]{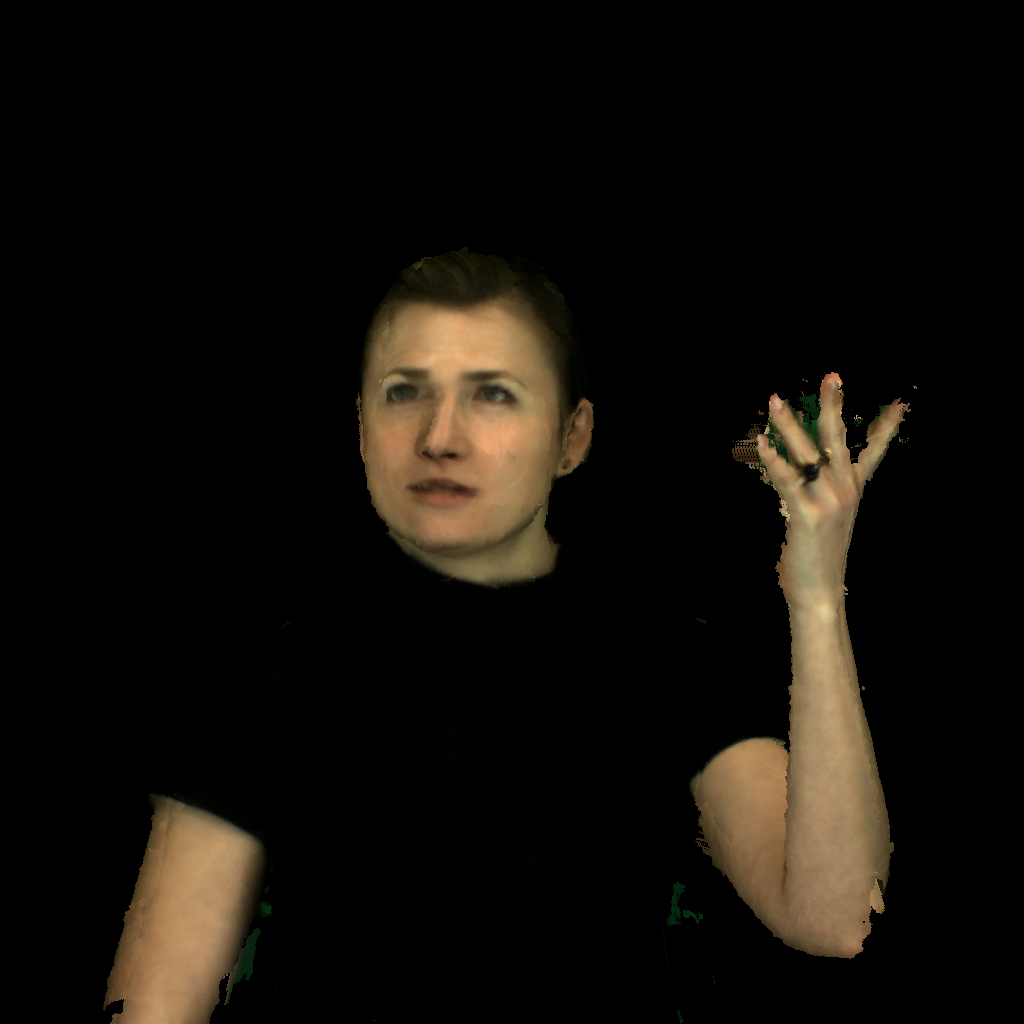}
        \caption{\textit{Sarah}}
    \end{subfigure}
    \hfill
    \begin{subfigure}[b]{0.15\textwidth}
        \includegraphics[width=\textwidth]{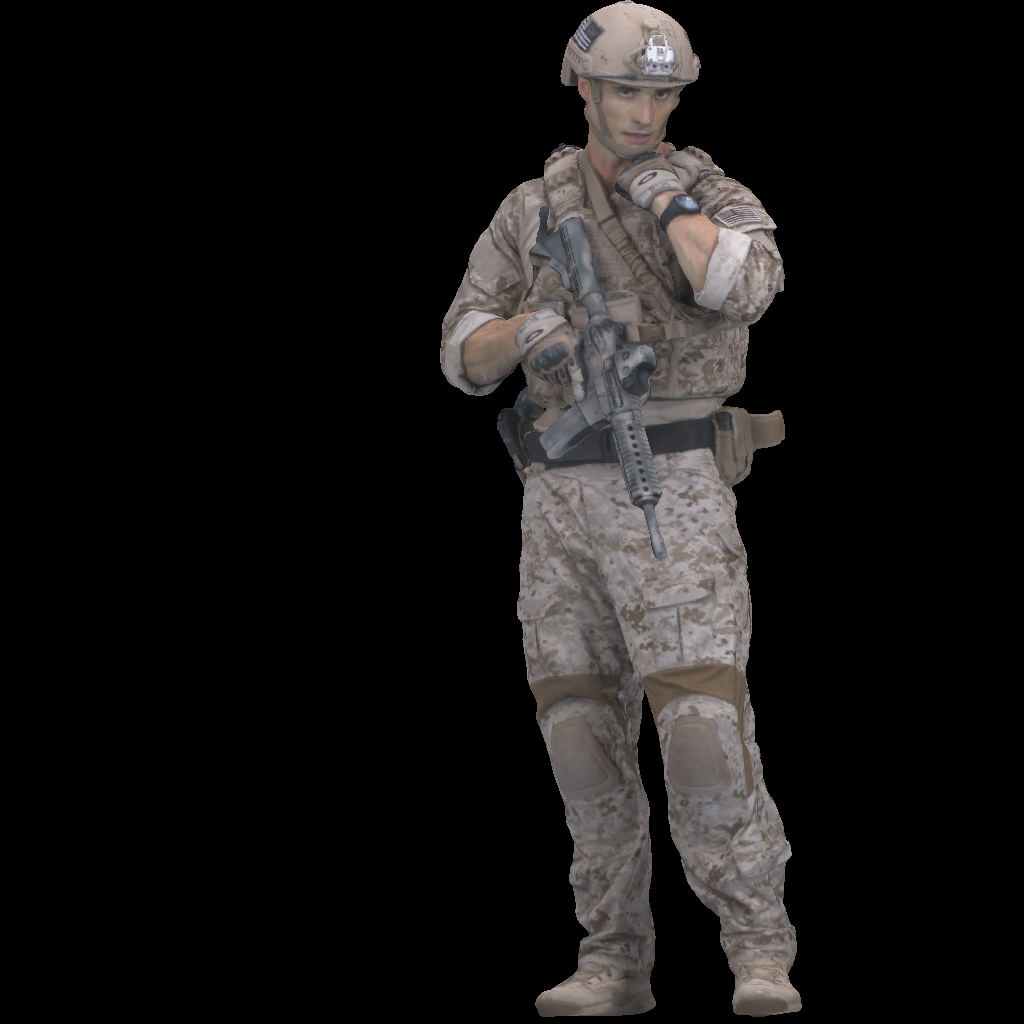}
        \caption{\textit{Soldier}}
    \end{subfigure}
    \captionsetup{font={small},justification=raggedright, singlelinecheck=off}
    \caption{The first frame of 9 point cloud sequences in testing dataset.}
    \label{fig:test dataset}
\end{figure}

\textbf{Testing Dataset.} We selected nine point cloud sequences from 8iVFBv2 \cite{8iVFBv2} and MVUB \cite{MVUB} as the testing dataset. Each point cloud sequence contains between 200 and 300 frames, with each frame consisting of 500,000 to 2,000,000 points, which contains rich texture information. 
%%%%%%%%%%%%%%%%%%%%%%%%%%%%%%arXiv
Table \ref{tab:dataset} presents the frame count of each point cloud sequence, while Figure \ref{fig:test dataset} illustrates the visual results of the first frame of each point cloud sequence.

\begin{table}
    \centering
    % \small
    \begin{tabular}{c|c|c}
        \toprule
        \hline
        \textbf{Dataset} & \textbf{Point Cloud Sequence} & \textbf{Frames} \\
        \cline{1-3}
        \multirow{5}{*}{\textbf{MVUB}} & \textit{Andrew} & 318\\
        \cline{2-3}
        & \textit{David} & 216 \\
        \cline{2-3}
        & \textit{Phil} & 245 \\
        \cline{2-3}
        & \textit{Ricardo} & 216 \\
        \cline{2-3}
        & \textit{Sarah} & 207 \\
        \cline{1-3}
        \multirow{4}{*}{\textbf{8iVFBv2}} & \textit{Longdress} & 300 \\
        \cline{2-3}
        & \textit{Loot} & 300 \\
        \cline{2-3}
        & \textit{Redandblack} & 300 \\
        \cline{2-3}
        & \textit{Soldier} & 300 \\
        \hline
        \bottomrule
    \end{tabular}
    \captionsetup{font={small}}
    \caption{The frame count of each point cloud sequence}
    \label{tab:dataset}
\end{table}

%%%%%%%%%%%%%%%%%%%%%%%%%%%%%%%%%%%%%%%%%%%arXiv

\subsection{Baseline Models}
\label{Baseline Models}
We apply the proposed method to the following VAE-based models to demonstrate the effectiveness of our method.

\textbf{SparsePCAC:} The first variational model with hyperprior and context for point cloud attribute compression, which generates autoregressive context serially using masked sparse convolutions \cite{wang2022sparse}.

\textbf{TSC-PCAC:} The TSCM module is introduced based on SparsePCAC, which combines transformer and sparse convolution to better capture both local and global information. Additionally, a channel-based context module is proposed to reduce encoding and decoding time \cite{guo2024tsc}.

\textbf{SPAC:} One of the current state-of-the-art frameworks based on VAE. SPAC proposes using Fast Fourier Transform to hierarchically process the point clouds and then compress them by SparsePCAC, where the high-frequency layers could leverage the contex from the low-frequency layers while coding \cite{mao2024spac}.

\subsection{Two-Step Training}
We apply the generalized Gaussian entropy model and dynamic likelihood intervals proposed in Section \ref{Proposed Method} to each baseline model in Section \ref{Baseline Models}.Then train these improved models and compare the coding results with those of the baseline models. Specifically, we replace the structures of the modules marked with * in Table \ref{tab:arch} (i.e., the modules in the classic variational compression model with hyperprior and context) with the corresponding module structures from baseline models. For the sake of simplicity and effectiveness in training, we perform following two-step training to prevent the model from being difficult to converge due to too many parameters:
\begin{itemize}
  \item First train the modules in the baseline models, including the autoencoder $(g_a,g_s)$, hyper autoencoder $(h_a,h_s)$, context model $g_{cm}$ and the estimating module $g_{ep}$ for mean and scale. Meanwhile, disable $h_{cm}, h_{ep}$ and $d_{me}$ we propose, and set $\bm{\beta}=1, \bm{\pi}=\frac{1}{2}$, thus the generalized Gaussian distribution degenerates into Laplacian distribution and dynamic likelihood intervals degenerates into $\mathcal{U}(-\frac{1}{2},\frac{1}{2})$:
\begin{equation}
\begin{split}
    p_{\bm{\hat{y}}|\bm{\hat{z}}}(\bm{\hat{y}}|\bm{\hat{z}})&=\prod_i \left[ \mathcal{G}(\mu_i, \sigma_i,1) * LI\left(\mu_i,\sigma_i,\frac{1}{2}\right) \right] (\hat{y}_i)\\
    &=\prod_i \left[ \mathcal{L}(\mu_i, \sigma_i) * \mathcal{U}\left(-\frac{1}{2},\frac{1}{2}\right) \right] (\hat{y}_i).
\end{split}
\end{equation}
  \item Train $h_{cm}, h_{ep}$ and $d_{me}$ while freezing the modules trained in the first step.
\end{itemize}

We set $\lambda=400,1000,4000,8000$ and 16000 in order to get different bitrate points.

\begin{table*}
    \centering
    \scalebox{0.75}{
    % \captionsetup{font={small}}
    % \caption{BD-BR (\%) comparisons of the proposed method with PLT and RAHT on 9 point cloud sequences.}
    %\Large
    \begin{tabular}{l|cc|cc|cc|cc|cc|cc}
        \toprule
        \hline
        \multirow{3}{*}{\textbf{Point Clouds}} & \multicolumn{4}{c|}{\textbf{Ours vs SparsePCAC}} & \multicolumn{4}{c|}{\textbf{Ours vs TSC-PCAC}} & \multicolumn{4}{c}{\textbf{Ours vs SPAC}} \\
        \cline{2-13}
                                    & \multicolumn{2}{c|}{BD-BR(\%)} & \multicolumn{2}{c|}{BD-PSNR(dB)} & \multicolumn{2}{c|}{BD-BR(\%)} & \multicolumn{2}{c|}{BD-PSNR(dB)} &
                                    \multicolumn{2}{c|}{BD-BR(\%)} &
                                    \multicolumn{2}{c}{BD-PSNR(dB)}\\
        \cline{2-13}
                                    & Y & YUV & Y & YUV & Y & YUV & Y & YUV & Y & YUV & Y & YUV\\
        \hline
        \textit{Andrew} & -10.46 & -11.97 & 0.35& 0.38 &-8.82 & -9.99 & 0.21 & 0.24 & -6.07 & -6.38 & 0.13 & 0.14\\
        \textit{David}  &-9.92& -11.93 &0.41& 0.46 &-9.37& -10.09 &0.31& 0.32 &-5.18& -5.35 &0.17& 0.17\\
        \textit{Longdress} &-10.39& -10.65 &0.34& 0.36 &-8.87& -9.31 &0.24& 0.25 &-5.29& -5.70 &0.15& 0.16\\
        \textit{Loot} &-10.83& -11.15 &0.52& 0.54 &-8.43& -9.67 &0.36& 0.39 &-4.29& -5.76 &0.18& 0.22\\
        \textit{Phil} &-9.35& -10.97 &0.39& 0.42 &-7.95& -9.40 &0.25& 0.29 &-5.03& -5.92 &0.13& 0.15\\
        \textit{Redandblack}   &-10.06& -10.50 &0.35& 0.36 &-8.02& -8.83 &0.24& 0.25 &-5.24& -5.87 &0.20& 0.20\\
        \textit{Ricardo} &-9.74& -11.07 &0.47& 0.49 &-8.76& -9.24 &0.33& 0.36 &-5.20& -5.93 &0.21& 0.22\\
        \textit{Sarah}  &-10.85& -11.00 &0.39& 0.42 &-8.34& -10.11 &0.27& 0.28 &-6.38& -8.13 &0.21& 0.24\\
        \textit{Soldier} &-11.96& -13.93 &0.59& 0.61 &-7.99& -8.91 &0.29& 0.32 &-4.77& -5.96 &0.18& 0.21\\
        \hline
        \textbf{Average} & \textbf{-10.40 } & \textbf{-11.46} & \textbf{0.42} & \textbf{0.45} & \textbf{-8.51} & \textbf{-9.51} & \textbf{0.28} & \textbf{0.30} & \textbf{-5.27} & \textbf{-6.11} & \textbf{ 0.17} & \textbf{ 0.19}\\
        \hline
        \bottomrule
    \end{tabular}}
    \captionsetup{font={small}, singlelinecheck=off}
    \caption{BD-BR and BD-PSNR comparisons between three baseline models improved by proposed method and original results.}
    \label{tab:bd-br}
\end{table*}
\begin{table}
    \centering
    \scalebox{0.63}{
    \small
    % \captionsetup{font={small}}
    % \caption{BD-BR (\%) comparisons of the proposed method with PLT and RAHT on 9 point cloud sequences.}
    %\Large
    \begin{tabular}{l|cc|cc}
        \toprule
        \hline
        \multirow{2}{*}{\textbf{Point Clouds}} & \multicolumn{2}{c|}{\textbf{SparsePCAC with GGEM}} & \multicolumn{2}{c}{\textbf{SparsePCAC with DLI}}  \\
        \cline{2-5}
                                    &{BD-BR(\%)} & {BD-PSNR(dB)} & {BD-BR(\%)} & {BD-PSNR(dB)} \\
                                    % &b & b & b &b \\
        \hline
        \textit{Andrew} & -5.94 &0.18& -6.51 & 0.30\\
\textit{David} & -5.03 &0.22& -6.60 & 0.28\\
\textit{Longdress} & -5.79 &0.16& -6.47 & 0.29 \\
\textit{Loot} & -5.19 &0.21& -7.70 & 0.30\\
\textit{Phil} & -6.08 &0.17& -5.49 & 0.25 \\
\textit{Redandblack} & -5.48 &0.20& -6.46 & 0.28 \\
\textit{Ricardo} & -5.82 &0.16& -6.16 & 0.28\\
\textit{Sarah} & -5.11 &0.18& -7.94 & 0.27 \\
\textit{Soldier} & -5.90 &0.19& -8.13 & 0.31\\
        \hline
        \textbf{Average} & \textbf{-5.59 } & \textbf{0.19} & \textbf{-6.83} & \textbf{0.28}\\
        \hline
        \bottomrule
    \end{tabular}}
    \captionsetup{font={small}, singlelinecheck=off}
    \caption{Ablation study for GGEM (generalized Gaussian entropy model) and DLI (dynamic likelihood interval) based on SparsePCAC.}
    \label{tab:ablation}
\end{table}

\subsection{Rate-Distortion Performance}
Figure \ref{fig:rd} illustrates the rate-distortion performance of different methods in nine point cloud sequences. % Bits per point (bpp) represent the bitrate, while Y-PSNR denotes the peak signal-to-noise ratio of Y component, which can be used as a measure of distortion between reconstructed point clouds and original point clouds.
Table \ref{tab:bd-br} shows the BD-BR and BD-PSNR gains achieved by applying generalized Gaussian entropy model and dynamic likelihood interval to baseline models. 
% BD-BR represents the difference in bitrate with the same distortion, while BD-PSNR represents the difference in distortion at the same bitrate.

As shown in Figure \ref{fig:rd} and Table \ref{tab:bd-br}, our method improves performance across all baseline models. The largest improvement is observed for SparsePCAC, which achieves an average bitrate reduction of 11.46\%, followed by TSC-PCAC with an average bitrate reduction of 9.51\%. The smallest improvement is observed for SPAC, which still results in an average bitrate saving of 6.11\%. We be believe this outcome can be attributed to the fact that SparsePCAC has the weakest performance among the baseline models, with less accurate entropy parameter estimation. Therefore, adjustments to the entropy model and likelihood intervals are more effective in this case. In contrast, SPAC already has relatively accurate estimation for mean and scale parameters, so even optimal adjustments lead to a smaller improvement in performance.

\begin{table}
    \centering
    \scalebox{0.95}{% 缩小到原来的80%
    \scriptsize 
    %\Large
    \begin{tabular}{l|cc|cc|cc}
        \toprule
        \hline
        \multirow{2}{*}{\textbf{Methos}} & \multicolumn{2}{c|}{\textbf{Ours vs SparsePCAC}} & \multicolumn{2}{c|}{\textbf{Ours vs TSC-PCAC}} & \multicolumn{2}{c}{\textbf{Ours vs SPAC}} \\
        \cline{2-7}
                                    & {Origin} & {Ours} & {Origin} & {Ours} &
                                    {Origin} &
                                    {Ours}\\
        \hline
         Enc.(s) & 74.05 & 163.74 & 3.92 & 8.24 & 132.60 & 287.59 \\
         Dec.(s) & 427.63 & 904.17 & 7.60 & 17.80 & 108.58 & 231.14 \\
        \hline
        \bottomrule
    \end{tabular}}
    \captionsetup{font={small}, singlelinecheck=off}
    \caption{Comparison of complexity. Take the average time of encoding and decoding all point clouds as results.}
    \label{tab:time}
\end{table}

%%%%%%%%%%%%%%%%%%%%%%%%%%%%%%%%arXiv
\subsection{Visual Results}
Figure \ref{fig:visual} shows the improvement of our method on the visual effect of SPAC in reconstructing point clouds.
%%%%%%%%%%%%%%%%%%%%%%%%%%%%%%%%arXiv

\subsection{Computational Complexity}
Table \ref{tab:time} illustrates the encoding and decoding time for compressing point cloud sequences using different methods.

Our method does not significantly affect the parallelism of baseline models' encoding and decoding, as we set the second context module $h_{cm}$ to be the same as $g_{cm}$. For TSC-PCAC with the highest parallelism, our method incurs the least additional encoding and decoding time. In contrast, for SparsePCAC, which uses a fully serial context module, our method incurs the most additional time.

\subsection{Ablation Study}
In order to separately demonstrate the effectiveness of generalized Gaussian entropy model and dynamic likelihood interval, we add only one of these two modules to SparsePCAC then perform two-step training, and the results are shown in Table \ref{tab:ablation}. We can learn from Table \ref{tab:ablation} that both generalized Gaussian entropy model and dynamic likelihood interval have a positive effect on improving coding performance, which is consistent with the results from the preliminary experiment.
\section{Conclusion and Future Work}

We conduct simple experiments to demonstrate the limitations of current Gaussian and Laplace entropy models. Then we propose using generalized Gaussian distribution for more accurate probability estimation of latents and leverage MED to dynamically adjust the likelihood intervals of integers. Our method significantly improves the RD performance of three tested VAE-based models. Since our method only adds a few modules to original models without fundamentally changing their architecture, and can be trained by proposed two-step training, it is easily portable to other SOTA VAE-based models.

In future work, we will focus on enhancing the parallelism of proposed modules and testing our method on other compression tasks, such as image and video compression.

% 首先利用简单的实验证明了。。。。的局限性，然后提出了。。。。性能。。。。由于时间未其他编码领域。。。并行的上下文

% 属性压缩。。。目前都是基于高斯拉普拉斯。。。实验证明高斯或拉普拉斯熵模型的参数还有可以利用的信息来改进上模型和似然区间。。。提出。。。性能。。。可扩展到整个深度编码领域
\newpage
{
    \newpage
    \small
    \bibliographystyle{ieeenat_fullname}
    \bibliography{main}
}


\end{document}